\documentclass[10pt,twocolumn,letterpaper]{article}

\usepackage[pagenumbers]{cvpr} %

\usepackage{multirow}
\usepackage{array}%

\newcommand{\strike}[1]{\begingroup\setbox0\hbox{#1}\rlap{\rule[0.55ex]{\wd0}{0.6pt}}\box0\endgroup}
\definecolor{figgreen}{RGB}{27,158,119}
\definecolor{figorange}{RGB}{217,95,2}
\definecolor{figgray}{RGB}{243,243,243}
\definecolor{figkey}{RGB}{176,120,0}   %

\definecolor{cvprblue}{rgb}{0.21,0.49,0.74}
\usepackage[pagebackref,breaklinks,colorlinks,allcolors=cvprblue]{hyperref}

\def\confName{CVPR}
\def\confYear{2027}

\title{\parbox[c]{2.3em}{\includegraphics[height=2.6em]{figs/logo}}\hspace{0.5em}%
  \begin{tabular}[c]{@{}c@{}}Causal Dioramas: Controllable Video Generation,\\Revision and Editing via Intervention\end{tabular}}

\author{%
Yifan Shen$^{1}$\quad Xinyi Yang$^{2}$\quad Peiyuan Zhu$^{1}$\quad Yichang Jian$^{3}$\quad Jinyuan Hu$^{1}$\\
Xilin He$^{1}$\quad Fengming Zhu$^{2}$\quad Guangyi Chen$^{1,3}$\quad Kun Zhang$^{3,1}$\\[0.4em]
{\normalsize $^{1}$Mohamed bin Zayed University of Artificial Intelligence\quad $^{2}$Tencent\quad $^{3}$Carnegie Mellon University}%
}

\makeatletter

\def\abstract{\iftoggle{cvprpagenumbers}{}{\thispagestyle{empty}}\centerline{\large\bf Abstract}\vspace*{12pt}\noindent\it\ignorespaces}
\makeatother

\makeatletter
\def\@maketitle{\newpage\null\vskip .1in\begin{center}{\Large \bf \@title \par}\vspace*{14pt}{\large\lineskip .5em
\begin{tabular}[t]{c}\@author\end{tabular}\par}\vskip .3em\vspace*{6pt}\end{center}}
\makeatother
\begin{document}
\twocolumn[{%
\renewcommand\twocolumn[1][]{#1}%
\maketitle
\newlength{\teaserinputh}\setlength{\teaserinputh}{2.67cm}
\newcommand{\teasertag}[2]{{\setlength{\fboxsep}{1.5pt}\colorbox{#1}{\fontsize{7}{7}\selectfont\sffamily\bfseries\color{white}\vphantom{Ag}#2}}}
\newcommand{\teaserstrip}[3]{%
  \sbox0{\setlength{\fboxsep}{0pt}\setlength{\fboxrule}{0.8pt}\fcolorbox{#1}{white}{\includegraphics[width=\dimexpr\linewidth-1.6pt\relax]{#3}}}%
  \sbox2{\teasertag{#1}{#2}}%
  \usebox0\llap{\makebox[\wd0][l]{\hspace{0.8pt}\raisebox{\dimexpr\ht0-0.8pt-\ht2\relax}{\usebox2}}}}
\newcommand{\teaserimg}[3][black]{\teaserstrip{#1}{#2}{#3}\\[1.5pt]}
\newcommand{\teaserempty}[2]{%
  \sbox0{\setlength{\fboxsep}{0pt}\setlength{\fboxrule}{0.8pt}\fcolorbox{gray!60}{figgray}{\parbox[c][\dimexpr(\linewidth-1.6pt)*360/1460\relax][c]{\dimexpr\linewidth-1.6pt\relax}{\centering\scriptsize\color{gray}#2}}}%
  \sbox2{\teasertag{gray!80}{#1}}%
  \usebox0\llap{\makebox[\wd0][l]{\hspace{0.8pt}\raisebox{\dimexpr\ht0-0.8pt-\ht2\relax}{\usebox2}}}\\[1.5pt]}
\definecolor{teaserarrow}{RGB}{64,64,64}
\newlength{\teaserlen}
\newcommand{\teaserharrow}[1]{\mbox{{\color{teaserarrow}\raisebox{-0.35pt}{\rule{\dimexpr#1-2.5pt\relax}{0.7pt}}}%
  \kern-0.3pt\raisebox{-1.45pt}{\includegraphics{figs/teaser_design/head_stealth}}}}%
\newcommand{\teaserdarrow}[1]{\mbox{\raisebox{-1.45pt}{\reflectbox{\includegraphics{figs/teaser_design/head_stealth}}}\kern-0.3pt%
  {\color{teaserarrow}\raisebox{-0.35pt}{\rule{\dimexpr#1-5pt\relax}{0.7pt}}}\kern-0.3pt\raisebox{-1.45pt}{\includegraphics{figs/teaser_design/head_stealth}}}}%
\newcommand{\teaservarrow}[2]{\ifx u#1\rotatebox{90}{\teaserharrow{#2}}\else
  \raisebox{#2}{\rotatebox{-90}{\teaserharrow{#2}}}\fi}%
\newlength{\teasergap}\setlength{\teasergap}{8pt}   %
\newlength{\teasermarkh}\setlength{\teasermarkh}{23pt}%
\newlength{\teaserarrl}\setlength{\teaserarrl}{8.5pt} %
\newlength{\teaserstripw}
\newif\ifteasershadow\teasershadowtrue
\newcommand{\teaserclip}[2]{%
  \sbox0{\setlength{\fboxsep}{0pt}\setlength{\fboxrule}{0.8pt}\fcolorbox{gray}{white}{\includegraphics[width=\dimexpr\linewidth-1.6pt\relax]{#2}}}%
  \sbox2{\teasertag{gray}{#1}}%
  \ifteasershadow\rlap{\kern-3pt\raisebox{\dimexpr-\dp0-3pt\relax}[0pt][0pt]{%
    \includegraphics[width=\dimexpr\wd0+6pt\relax,height=\dimexpr\ht0+\dp0+6pt\relax]{figs/teaser_design/clip_shadow}}}\fi
  \usebox0\llap{\makebox[\wd0][l]{\hspace{0.8pt}\raisebox{\dimexpr\ht0-0.8pt-\ht2\relax}{\usebox2}}}%
  \llap{\raisebox{\dimexpr\ht0-7.2pt\relax}[0pt][0pt]{%
    \includegraphics{figs/teaser_design/clip_dogear\ifteasershadow\else_flat\fi}}\kern-0.6pt}}
\newif\ifteaserdo
\newcommand{\teaserdoarrow}[1]{%
  \llap{\raisebox{\dimexpr#1/2-0.55ex\relax}{\color{figorange}\fontsize{7}{7}\selectfont do\kern1.8pt}}%
  \raisebox{#1}{\rotatebox{-90}{\mbox{{\color{figorange}\raisebox{-0.35pt}{\rule{\dimexpr#1-2.5pt\relax}{0.7pt}}}%
    \kern-0.3pt\raisebox{-1.45pt}{\includegraphics{figs/teaser_design/head_stealth_orange}}}}}}
\newcommand{\teasertoprow}[4]{%
  \sbox4{\includegraphics[height=\teasermarkh]{#4}}%
  \setlength{\teaserstripw}{\dimexpr\linewidth-\wd4-\teaserarrl-7pt\relax}%
  \sbox6{\setlength{\linewidth}{\teaserstripw}\teaserclip{#1}{#2}}%
  \setlength{\teaserlen}{\dimexpr(\ht6-\ht4)/2+\teasergap-1.4pt\relax}%
  \hbox to \linewidth{\usebox6\hfill
    \raisebox{\dimexpr\ht6/2\relax}{\ifx r#3\teaserharrow{\teaserarrl}\else\ifx e#3\makebox[\teaserarrl][c]{\color{teaserarrow}\rlap{\raisebox{0.75pt}{\rule{5.6pt}{0.7pt}}}\raisebox{-1.45pt}{\rule{5.6pt}{0.7pt}}}\else\makebox[\teaserarrl][c]{\color{teaserarrow}\rlap{\raisebox{-0.35pt}{\rule{6pt}{0.7pt}}}\kern2.65pt\raisebox{-3.35pt}{\rule{0.7pt}{6pt}}\kern2.65pt}\fi\fi}\hfill
    \raisebox{\dimexpr(\ht6-\ht4)/2\relax}{\usebox4\llap{\makebox[\wd4][c]{\raisebox{\dimexpr\ht4+0.6pt\relax}[0pt][0pt]{\ifteaserdo\teaserdoarrow{\teaserlen}\else\teaservarrow{d}{\teaserlen}\fi}}}}%
    \kern0.4pt}}
\newcommand{\teaserkey}[1]{\textcolor{figkey}{\textbf{#1}}}%
\newcommand{\teaserbox}[1]{{\ifteaserdo\colorlet{teaseracc}{figorange}\else\colorlet{teaseracc}{teaserarrow}\fi
  \sbox0{\colorbox{figgray}{\parbox[c][26pt][c]{\dimexpr\linewidth-2\fboxsep\relax}{\centering\footnotesize\itshape #1}}}%
  \usebox0\llap{\makebox[\wd0][l]{\raisebox{-\dp0}{\color{teaseracc}\rule{1.4pt}{\dimexpr\ht0+\dp0\relax}}}}}}
\newcommand{\teaserinputarea}[2]{\parbox[t][\teaserinputh][t]{\linewidth}{\noindent#2\par\nointerlineskip\vskip\teasergap\noindent#1}\\}
\newcommand{\teaserfaderow}[4]{%
  \sbox6{\hspace{0.8pt}\includegraphics[width=\dimexpr\linewidth-1.6pt\relax]{#2}\hspace{0.8pt}}%
  \sbox2{\teasertag{gray}{#1}}%
  \sbox4{\includegraphics[height=15pt]{#4}}%
  \setlength{\teaserlen}{\dimexpr(\ht6-\ht4)/2+\teasergap-1.1pt\relax}%
  \usebox6\llap{\makebox[\wd6][l]{\hspace{0.8pt}\raisebox{\dimexpr\ht6-\ht2\relax}{\usebox2}}}%
  \llap{\raisebox{\dimexpr(\ht6-\ht4)/2\relax}{%
    \ifx p#3\else
    \ifx r#3\raisebox{\dimexpr\ht4/2\relax}{\teaserharrow{8pt}}\else
    \raisebox{\dimexpr\ht4/2-0.55ex\relax}{\makebox[6.5pt][c]{\color{teaserarrow}\footnotesize$\equiv$}}\fi\fi
    \kern1pt\usebox4\llap{\makebox[\wd4][c]{\raisebox{\dimexpr\ht4+1.1pt\relax}[0pt][0pt]{\teaservarrow{d}{\teaserlen}}}}\kern1pt}}}
\newcommand{\teasertextcd}[3]{%
  \sbox4{\includegraphics[height=16pt]{#2}}%
  \setlength{\teaserstripw}{\dimexpr\linewidth-\wd4-\teaserarrl-3pt\relax}%
  \sbox6{\raisebox{\depth}{{\setlength{\fboxsep}{2pt}\colorbox{figgray}{\parbox[c][25.5pt][c]{\dimexpr\teaserstripw-4pt\relax}{\centering\fontsize{8}{8.4}\selectfont\itshape #1}}}}}%
  \setlength{\teaserlen}{\dimexpr(\ht6-\ht4)/2+\teasergap-1.4pt\relax}%
  \hbox to \linewidth{\usebox6\hfill\raisebox{\dimexpr\ht6/2\relax}{\teaserharrow{\teaserarrl}}\hfill
    \raisebox{\dimexpr(\ht6-\ht4)/2\relax}{\usebox4\llap{\makebox[\wd4][c]{\raisebox{\dimexpr\ht4+1.1pt\relax}[0pt][0pt]{%
      \ifx d#3\teaservarrow{d}{\teaserlen}\else\ifx e#3{\color{teaserarrow}\rule{0.6pt}{\teaserlen}\kern1.3pt\rule{0.6pt}{\teaserlen}}\fi\fi}}}}%
    \kern1pt}}
\newcommand{\teaserinputareaC}[2]{\parbox[t][\teaserinputh][t]{\linewidth}{\noindent#1\par\nointerlineskip\vskip\teasergap\noindent#2}\\}
\newcommand{\teasersep}{\makebox[\linewidth]{\raisebox{-2.1pt}[0.6pt][0pt]{\includegraphics[width=\linewidth]{figs/teaser_design/flow_connector}}}\\[2pt]}
\begin{center}
  \vspace*{0pt}%
  \captionsetup{type=figure}
  \begin{minipage}[t]{0.328\textwidth}
    \centering{\small\bfseries (a) Generation}\\[2pt]
    \teaserinputarea{\teasertoprow{Noise}{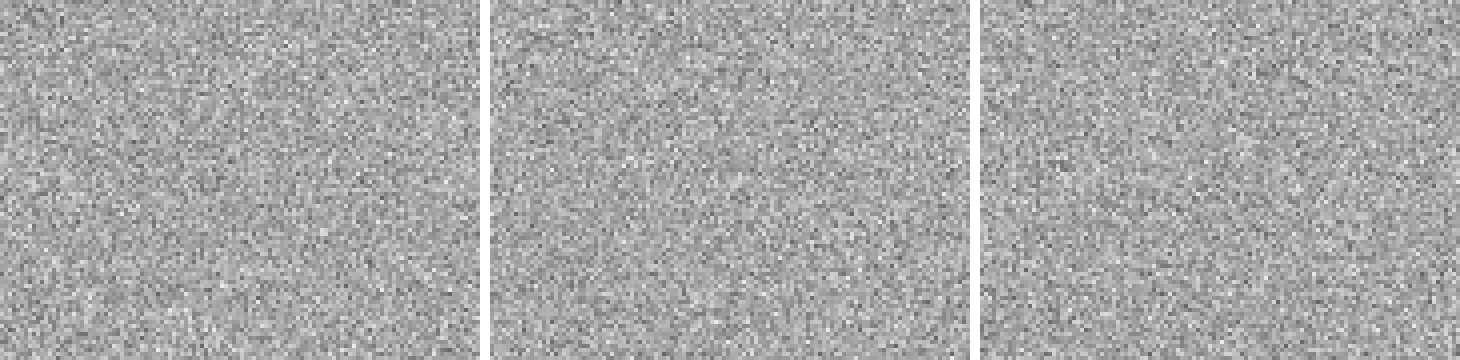}{p}{figs/teaser_design/cd_card_green}}
      {\teaserbox{``A hot air balloon\\floating \teaserkey{under} a sheep''}}
    \teasersep
    \teaserimg[figgreen]{Ours}{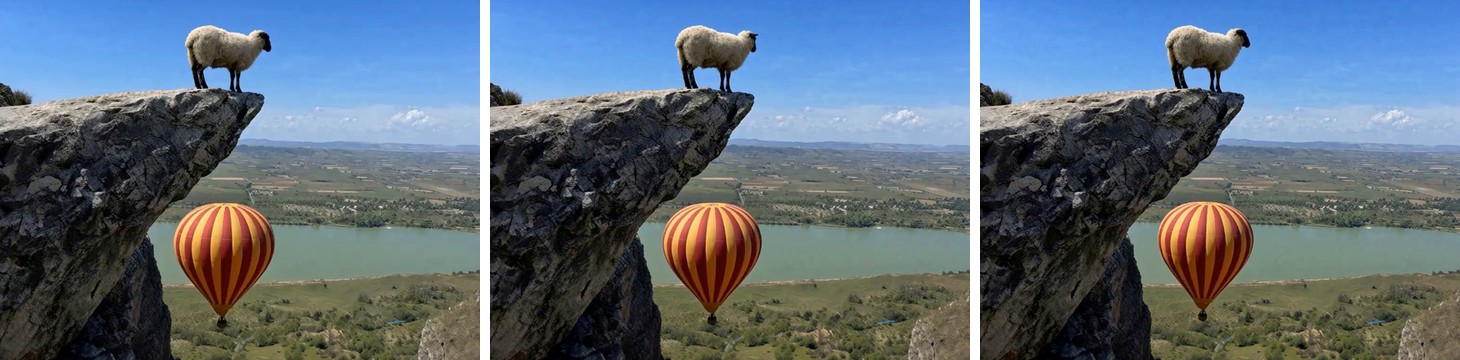}
    \teaserimg{PixVerse V3}{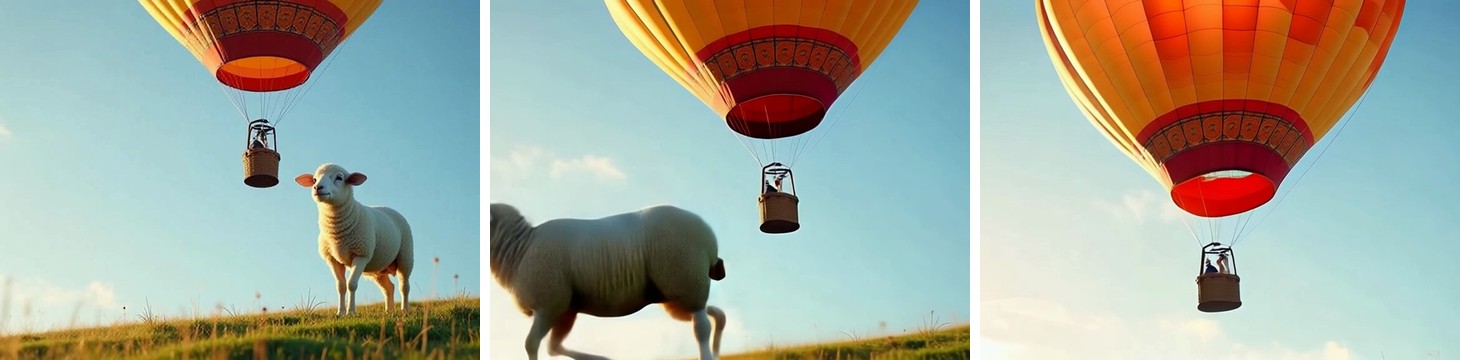}
    \teaserimg{Kling 1.0}{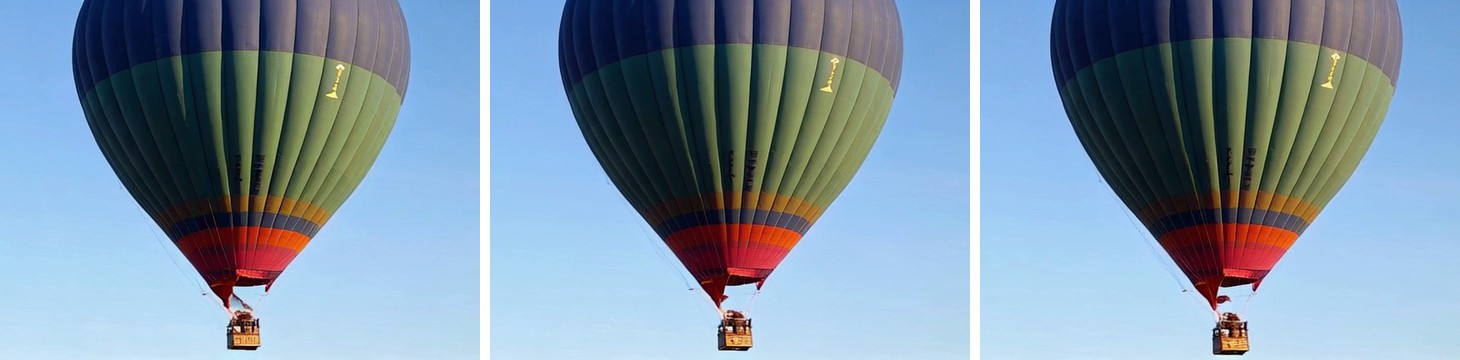}
    \teaserimg{Dreamina 1.2}{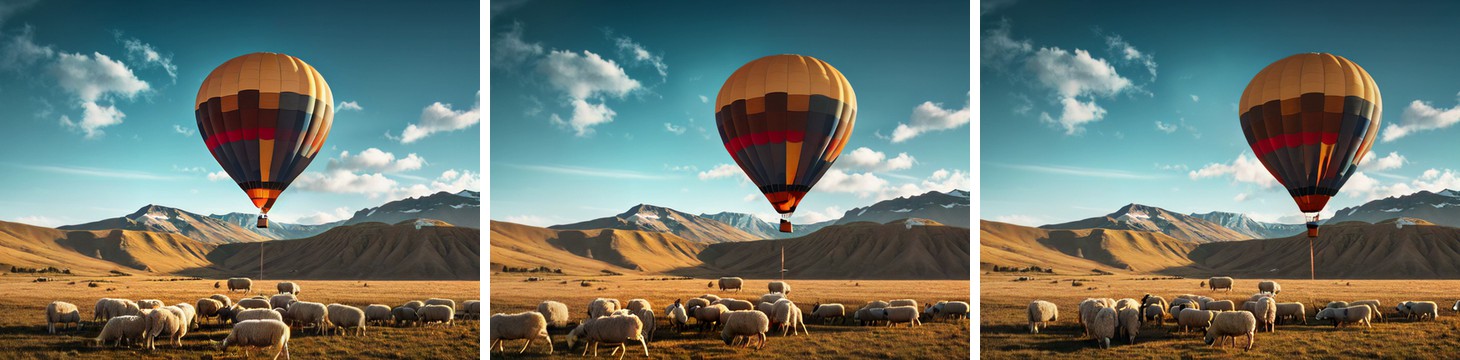}
    \teaserimg{CogVideoX 5B}{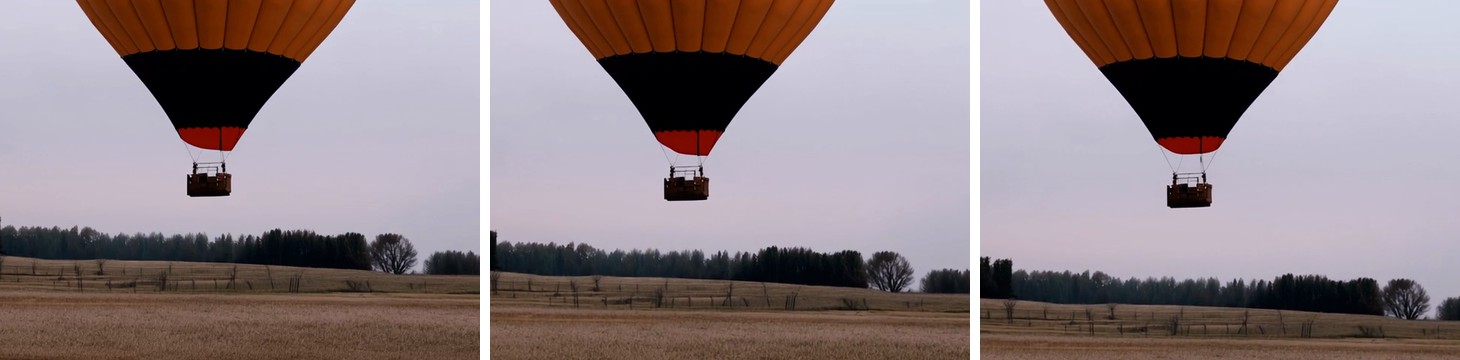}
  \end{minipage}\hfill
  \begin{minipage}[t]{0.328\textwidth}\teaserdotrue
    \centering{\small\bfseries (b) Revision}\\[2pt]
    \teaserinputarea{\teasertoprow{Source}{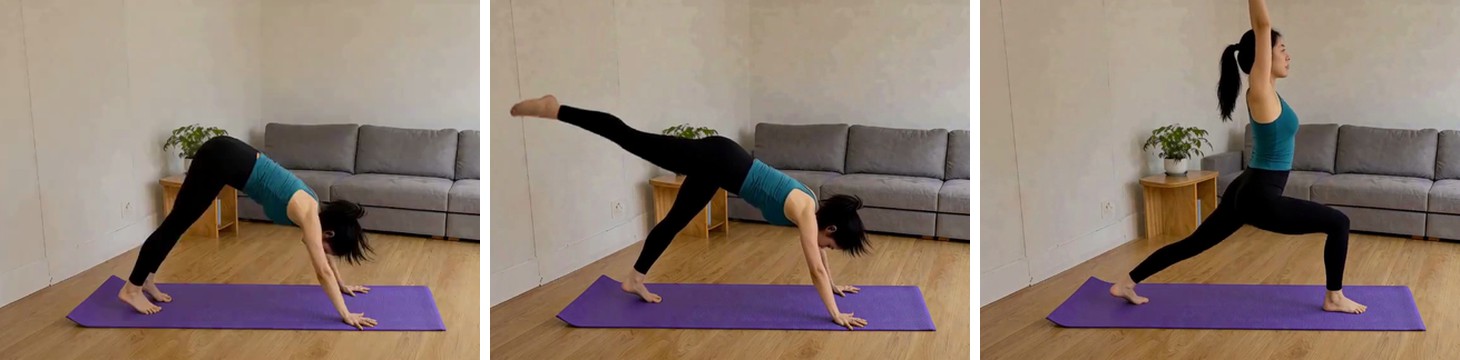}{e}{figs/teaser_design/cd_card_color}}
      {\teaserbox{``Change her yoga sequence to \teaserkey{sitting}\\cross-legged and \teaserkey{reading} a book.''}}
    \teasersep
    \teaserimg[figgreen]{Ours}{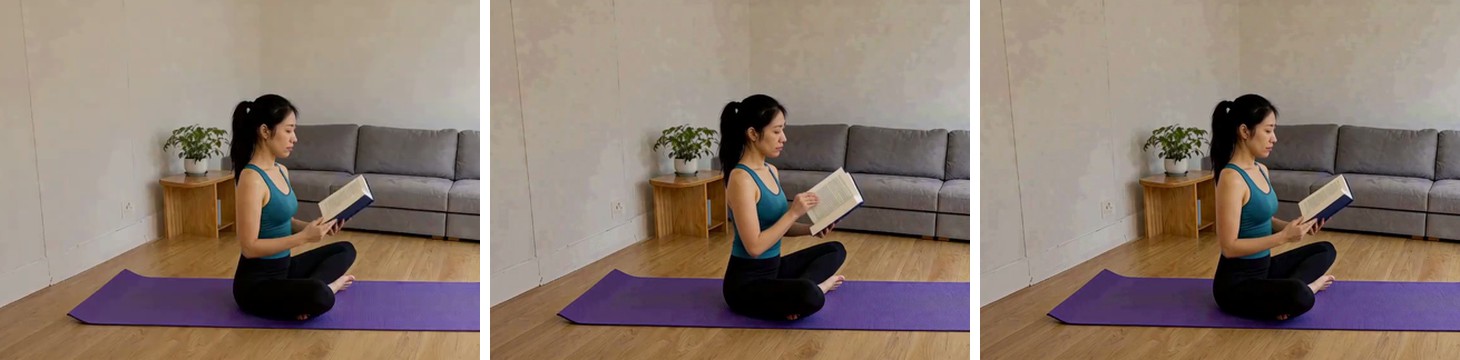}
    \teaserimg{OmniWeaving}{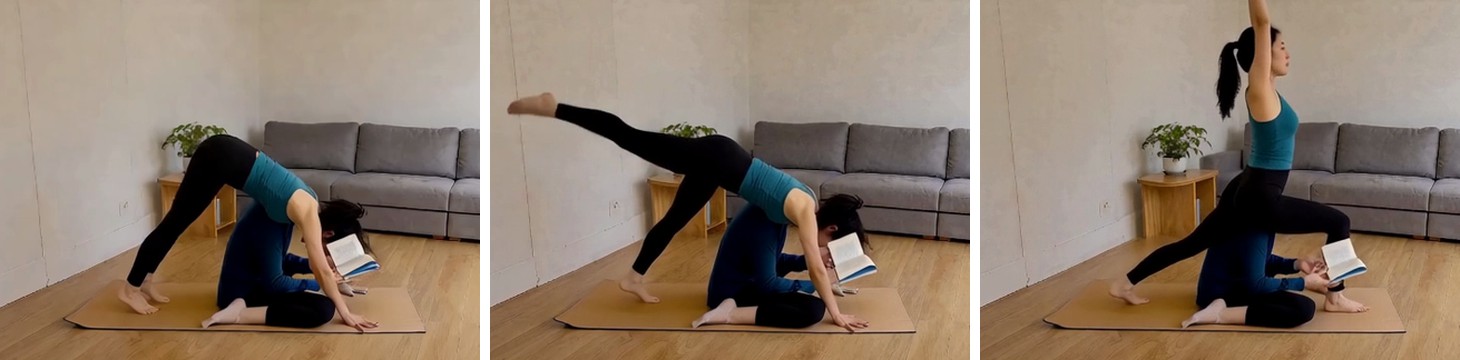}
    \teaserimg{Ditto}{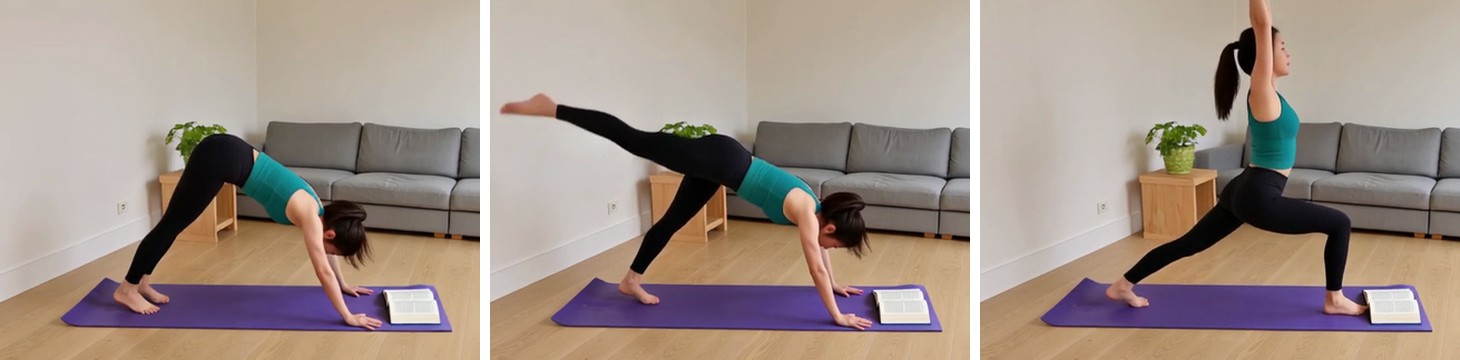}
    \teaserimg{Kiwi}{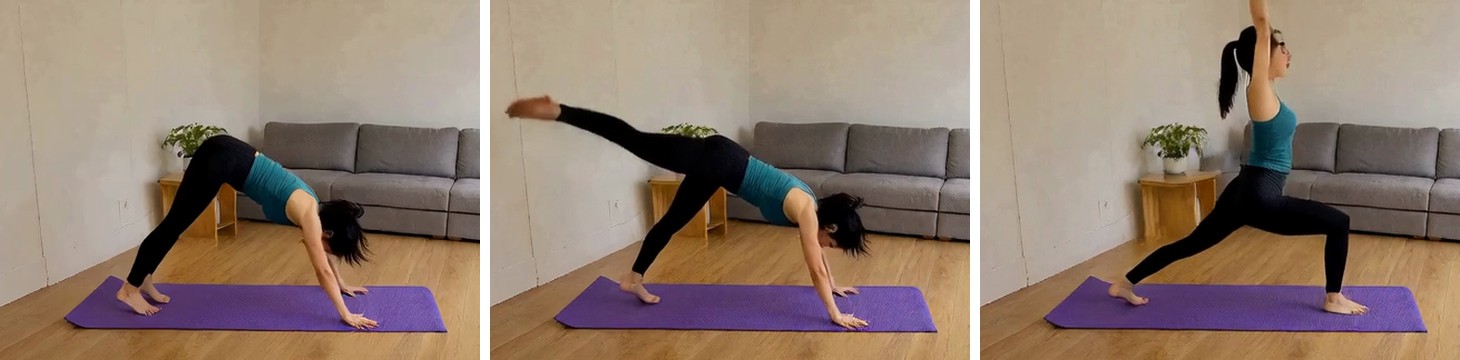}
    \teaserimg{Lucy}{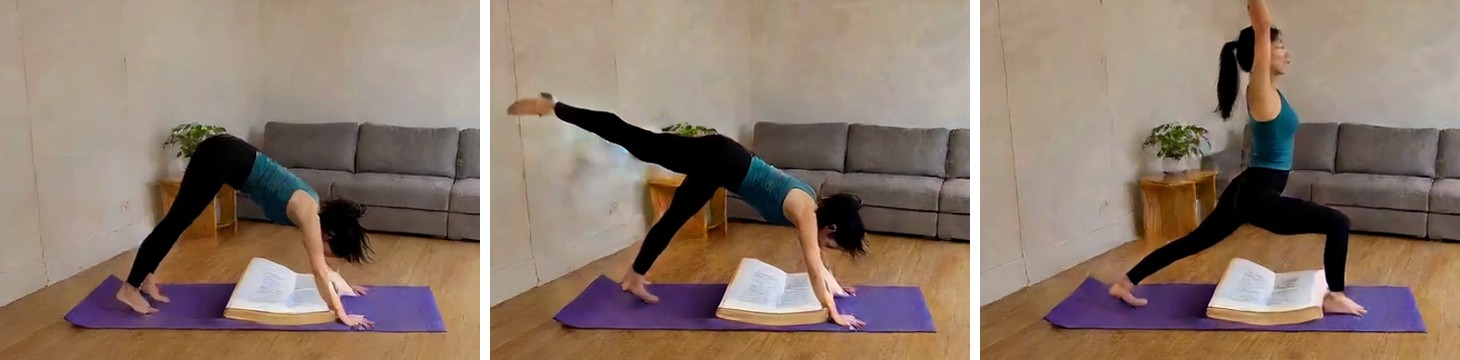}
  \end{minipage}\hfill
  \begin{minipage}[t]{0.328\textwidth}\teaserdotrue
    \centering{\small\bfseries (c) Editing}\\[2pt]
    \teaserinputarea{\teasertoprow{Source}{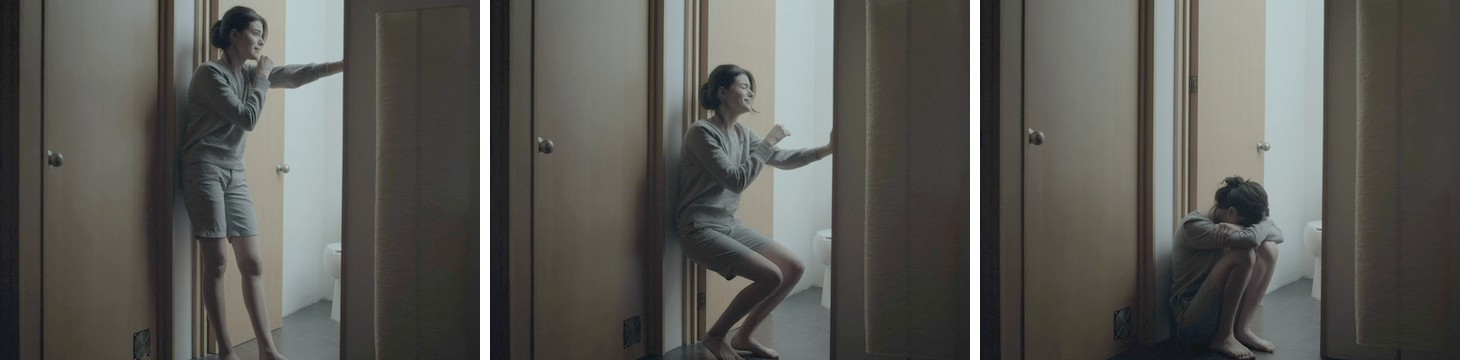}{r}{figs/teaser_design/cd_card_gray_do}}
      {\teaserbox{``Change her outfit to a \teaserkey{sweater and trousers},\\and let her \teaserkey{stand} leaning on the door frame.''}}
    \teasersep
    \teaserimg[figgreen]{Ours}{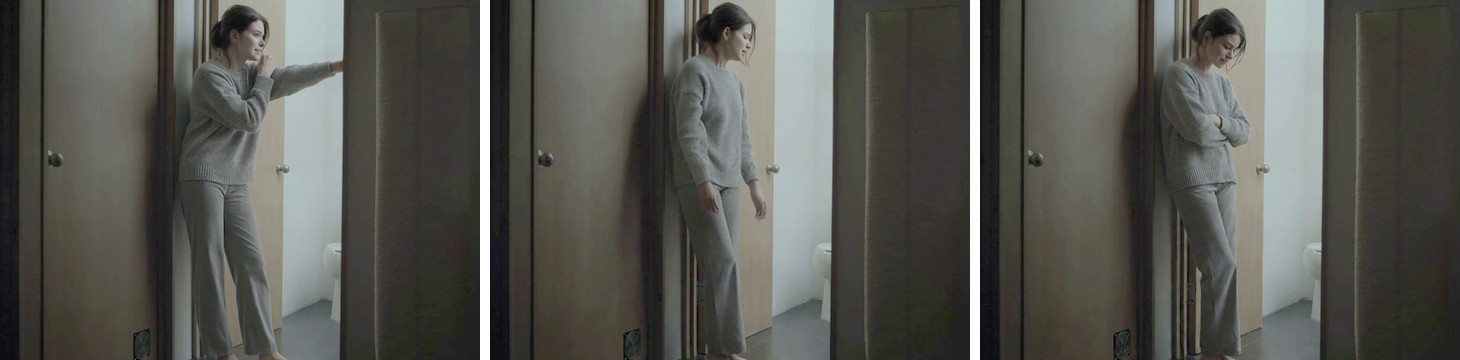}
    \teaserimg{Seedance 2.0}{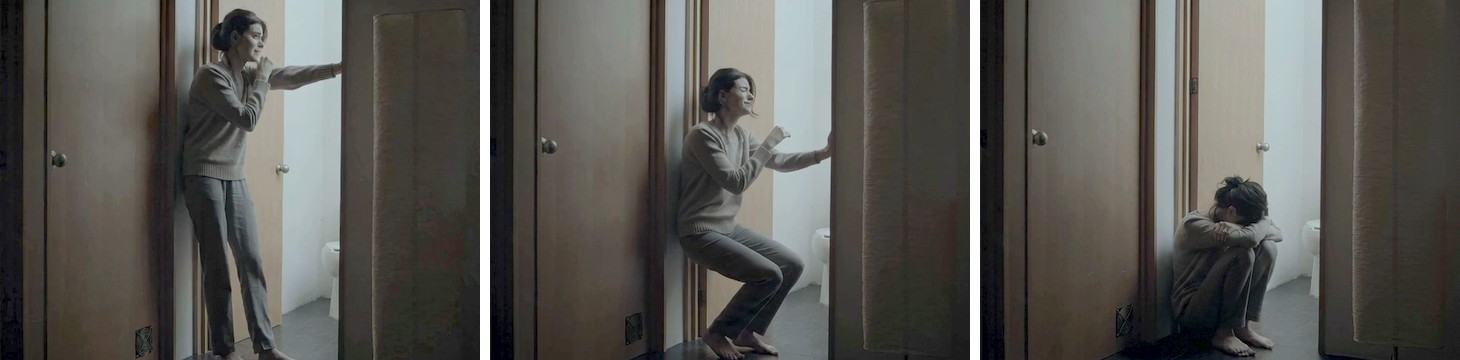}
    \teaserimg{Kling 3.0 Omni}{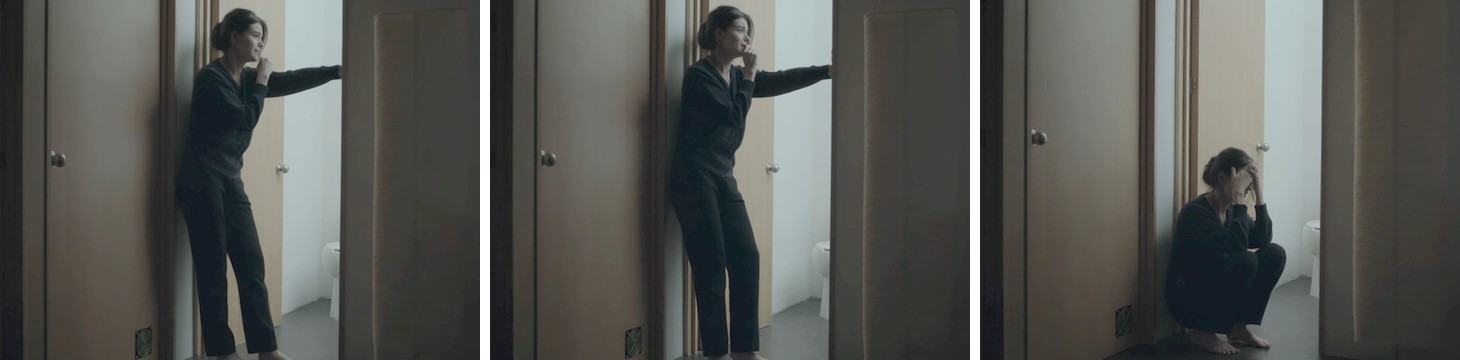}
    \teaserimg{Bernini}{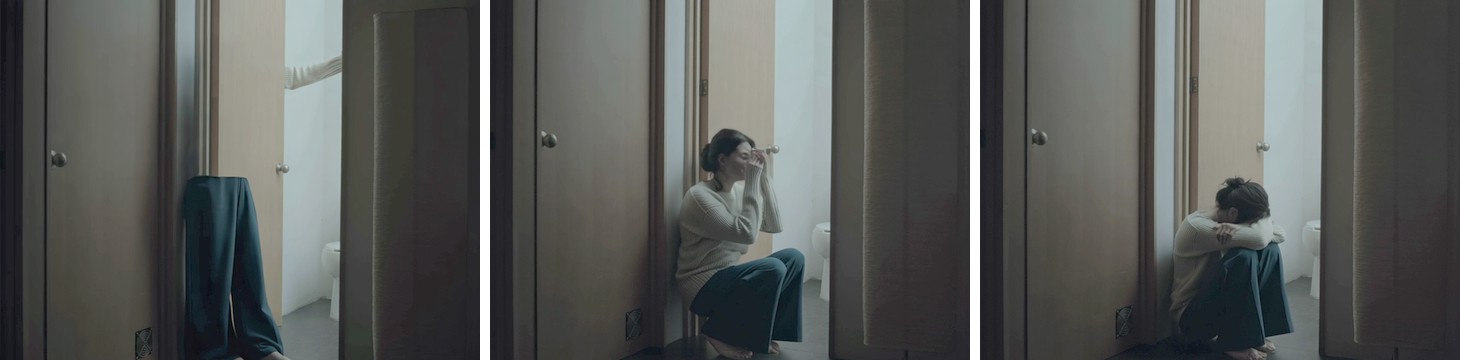}
    \teaserimg{OmniWeaving}{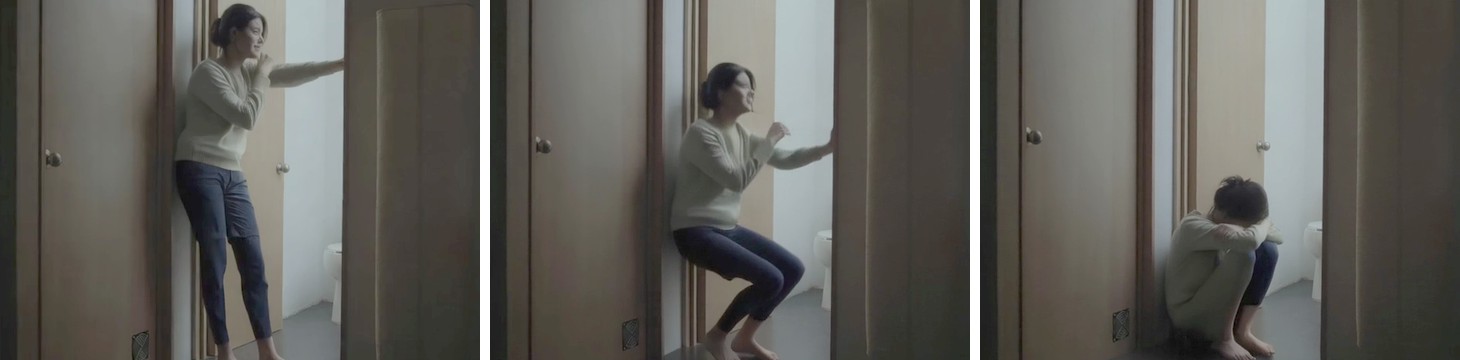}
  \end{minipage}
  \captionof{figure}{\textbf{Comparisons between the Causal Diorama and others on generation, revision and editing.}
    Generation builds the Causal Diorama from the prompt and renders it under sampled noise (+), whereas revision and editing intervene on it with an instruction (do).
    The Causal Diorama of its generated video is known exactly (=), while that of a real video has to be inferred ($\rightarrow$).
    Our method completes every task and preserves the unaffected content, while the other methods leave the task incomplete or introduce unintended changes.}
  \label{fig:teaser}
\end{center}

}]
\begin{abstract}
Two centuries ago, an early cinema-like theater called the Diorama could already control its scenes at will.
Today's video generation models produce videos of near-cinematic quality, yet they still lack this controllability.
This limitation stems from the inherent semantic gap between text and pixels, since a prompt can describe a video only coarsely and leaves its details and relations beyond control.
We bring this ability to video generation with Causal Dioramas, modules that decompose a video into a concept graph of variables that can be intervened on individually.
Each instruction is translated into an intervention on the concept graph, modifying the specified concepts and their dependents. A connector then conveys this intervention to the video generator, which reuses the original noise to produce a counterfactual video while preserving all other content.
Building on Causal Dioramas, we develop a unified framework for controllable generation, counterfactual revision of generated videos and editing of real videos.
We outperform both open-source and commercial models in controllability on T2V-CompBench, CoVEBench and OpenVE-Bench.
\end{abstract}

\section{Introduction}
\label{sec:intro}
\vspace{-4pt}

In 1822, Louis Daguerre and Charles Marie Bouton opened the Diorama in Paris, a theater often regarded as a precursor of cinema.
Operators could change its scenes at will through an intermediate device instead of repainting the canvas~\cite{whissel2007precinema}.
Two centuries later, video generation models can already produce videos of near-cinematic quality~\cite{seedance2026v2,kling2025omni,minimax2026h3,wu2025hunyuanvideo15,wan2025wan}.
Yet controllability remains a challenge, as these models still lack a modern \emph{Diorama} that reliably turns text into the intended video.

\begin{figure}[t]
  \centering
  \includegraphics[width=0.497\linewidth]{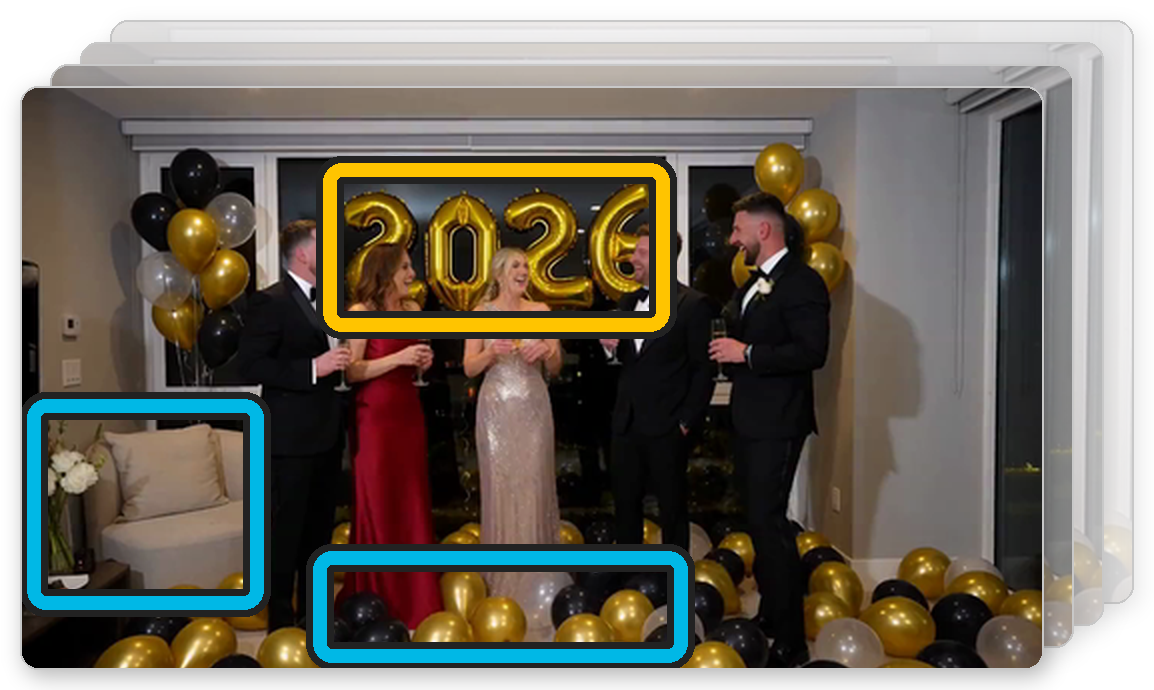}\hfill
  \includegraphics[width=0.497\linewidth]{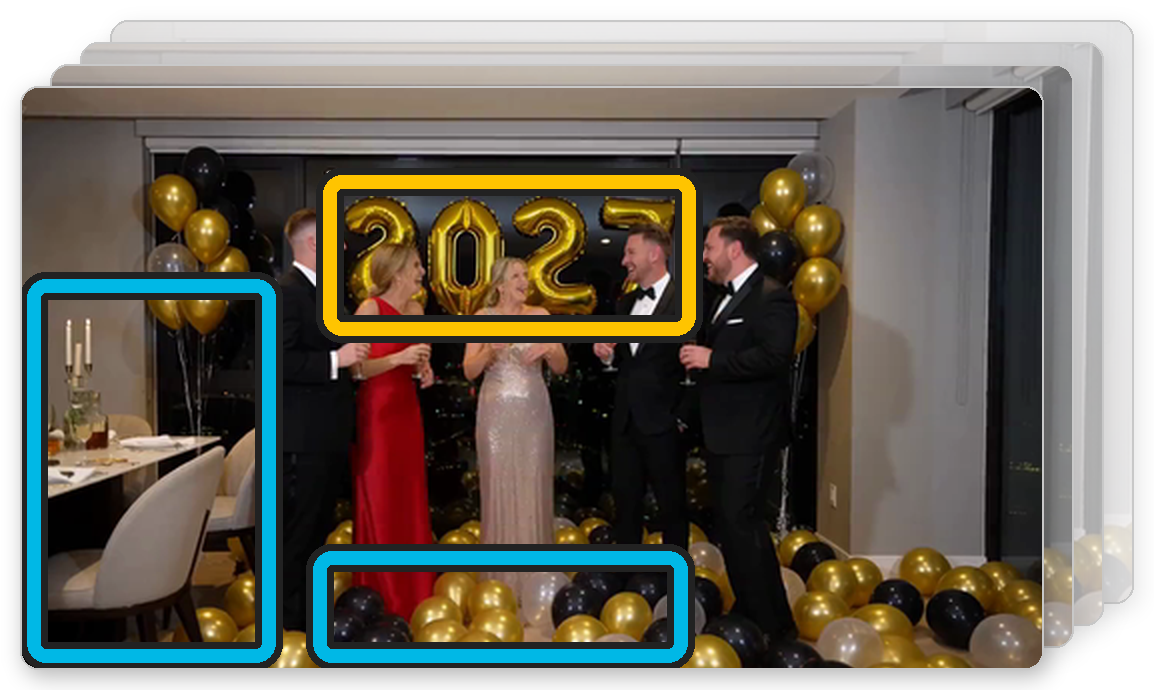}\\[1pt]
  \colorbox{figgray}{\parbox{0.97\linewidth}{\centering\footnotesize
    ``\ldots\ gold foil balloons spelling \strike{`2026'}\,\textcolor{figkey}{\textbf{`2027'}} are visible \ldots''}}
  \vspace{-6pt}
  \caption{\textbf{Changing a single word of the prompt.}
    Both videos are generated with the same noise and prompts that differ in only one word.
    Apart from the requested change (yellow), the sofa and the balloons on the ground also change unintentionally (blue).}
  \label{fig:oneword}
  \vspace{-16pt}
\end{figure}

We attribute this limitation to the inherent semantic gap between text and pixels.
A prompt describes a video only coarsely and entangles all of its concepts in a single sequence of text, whereas the video must render each of them at a specific place and time.
As a result, these models often omit described objects~\cite{zhang2026compattn}, produce the wrong number of instances~\cite{sun2026numina}, misplace objects relative to one another~\cite{liu2026spatialalign} and render incoherent motion~\cite{chefer2025videojam}.
Editing exposes the problem most clearly, since a faithful edit is an intervention that changes only the concepts specified by the instruction and those that depend on them~\cite{pearl2016primer}.
Such an edit asks a counterfactual question about this very video, which the joint distribution of prompts and videos cannot answer~\cite{bareinboim2022hierarchy}.
As shown in \Cref{fig:oneword}, changing a single word in the prompt can alter the entire video even with the generation noise fixed~\cite{hertz2023p2p}.

Existing methods for both generation and editing narrow this semantic gap from different directions.
Text-side methods rewrite the prompt in greater detail~\cite{yang2026scmapr,gao2025rapo,cheng2025vpo,ji2024promptavideo}, yet all concepts remain entangled within a single prompt.
Pixel-side methods condition the generator on visual signals such as depth maps, sketches, trajectories and camera poses for generation~\cite{wang2023videocomposer,guo2024sparsectrl,wang2024motionctrl,he2025cameractrl} or on features inverted from the source video for editing~\cite{qi2023fatezero,geyer2024tokenflow,ku2024anyv2v}, thereby achieving precise control over pixels rather than concepts.
Hybrid methods either learn the alignment between text and pixels implicitly from paired data, mostly for editing~\cite{cheng2024insv2v,bai2025ditto,lin2026kiwiedit,pan2026omniweaving,jiang2025vace}, or steer generation with coarse visual guides produced by a language model~\cite{lian2024lvd,lin2024videodirectorgpt,lv2023gpt4motion,li2025videomsg,chen2026cwm}.
Although these methods improve controllability, they do not separate the concepts of a video into variables that can be intervened on individually.
As a result, their performance remains limited on complex compositions and on edits that should preserve unrelated content.

We propose the \emph{Causal Diorama}, a module that decomposes a video into concept variables that can be intervened on individually.
Like Daguerre's Diorama, it stands between the prompt and the video, and each video is generated and modified through it.
Each concept is a separate variable that holds its own details such as appearance, position and motion, so the gap between text and pixels is bridged concept by concept.
It also records how these concepts depend on one another.
These variables and their dependencies form its \emph{concept graph}, and a fixed \emph{connector} turns the graph into conditions that the video generator understands.
We view the Causal Diorama and the video generator through the lens of structural causal models~\cite{pearl2016primer}, treating the generation noise as the exogenous noise of the video.
Under this view, each instruction becomes an intervention on the concept graph, which updates the specified concepts and the concepts that depend on them while leaving all others unchanged.
Regenerating with the same exogenous noise yields the counterfactual of the original video.
In this way, Causal Dioramas help video generation models follow the prompt or instruction while preserving the content it does not mention.

We validate Causal Dioramas on three tasks, namely controllable generation, counterfactual revision of generated videos and editing of real videos.
We introduce revision as a new task that reflects the common workflow of refining a generated video with a follow-up instruction.
The source video in revision is generated from a Causal Diorama with recorded noise, so each instruction intervenes directly on this Causal Diorama.
Editing applies the same interventions to a real video, whose Causal Diorama and exogenous noise must first be inferred from the video itself by abduction.
On T2V-CompBench~\cite{sun2024t2vcompbench}, our controllable generation surpasses both open-source and commercial video generators as well as the methods built on them.
In both revision and editing, we outperform existing editing models by a large margin on CoVEBench~\cite{wu2026covebench}, whose complex instructions often combine several changes.
Our editing also achieves the best overall score on the simpler single-type instructions of OpenVE-Bench~\cite{he2026openve}.

Our main contributions are summarized as follows.
\begin{itemize}[leftmargin=*,itemsep=2pt,topsep=2pt]
  \item We propose the Causal Diorama, a module decomposing a video into concept variables that can be intervened on individually and connecting them to a video generator.
  \item Building on the Causal Diorama, we develop a unified framework for controllable generation, counterfactual revision of generated videos and editing of real videos.
  \item We outperform both open-source and commercial models in controllability on T2V-CompBench, CoVEBench and OpenVE-Bench.
\end{itemize}

\section{Related Work}
\label{sec:related}

\noindent\textbf{Controllable video generation.}
Open-source~\cite{yang2024cogvideox,wan2025wan,wu2025hunyuanvideo15,hacohen2026ltx2,chen2024videocrafter2,li2024t2vturbov2,genmo2024mochi,minimax2026h3} and commercial~\cite{kuaishou2024kling,pixverse2024,capcut2024dreamina,runway2024gen3,pika2024,google2025veo,seedance2026v2} video generation models now produce near-cinematic videos, yet their controllability remains limited by the gap between text and pixels.
Text-side methods rewrite the prompt by enriching it with details~\cite{yang2026scmapr,jia2026cape}, tailoring it to the generator~\cite{gao2025rapo,cheng2025vpo,ji2024promptavideo} or correcting it after generation~\cite{song2026vqqa,choi2025neuse}.
Pixel-side methods condition the generator on visual signals that specify structure through depth maps and sketches~\cite{wang2023videocomposer,guo2024sparsectrl} or motion through trajectories and camera poses~\cite{wang2024motionctrl,he2025cameractrl}.
In addition, hybrid methods either steer attention inside the generator~\cite{yang2024vico,tian2024videotetris,zhang2026compattn,qu2025ttom,shaulov2026cvg} or provide it with coarse visual guides from a language model~\cite{lian2024lvd,lin2024videodirectorgpt,lv2023gpt4motion,li2025videomsg,wang2024dreamrunner,huang2024genmac,he2025dystxl,hassan2025avg,chen2026cwm,li2026videococo}.
However, the concepts of a video remain entangled in all these methods and cannot be intervened on individually.

\noindent\textbf{Video editing.}
Instruction-based methods formulate editing as video generation conditioned on the source video and the instruction.
These range from dedicated editing models~\cite{cheng2024insv2v,wu2025insvie,liao2025icve,zhang2025reco,bai2025ditto,decart2025lucyedit,lin2026kiwiedit,he2026openve,motamed2026void} to open-source~\cite{jiang2025vace,pan2026omniweaving,wei2025univideo,tan2025omnivideo,chen2026vino,shi2026mamoda25,bernini2026} and commercial~\cite{kling2025omni,runway2025aleph,happyhorse2026,seedance2026v2} unified models that handle both generation and editing.
Training-free methods instead reuse a pretrained generator, either inverting the source video and modifying its features~\cite{qi2023fatezero,geyer2024tokenflow,ku2024anyv2v,rout2024rfinversion} or mapping it directly to the target along the sampling path of a flow model without inversion~\cite{kulikov2025flowedit,kong2025ifv2v}.
A simpler alternative edits the prompt and regenerates the video with shared noise and attention~\cite{hertz2023p2p,liu2023videop2p}.
These methods edit the video as a whole, so a change often spreads beyond what the instruction specifies, while the Causal Diorama confines it to the specified concepts and the concepts that depend on them.

\noindent\textbf{Causal models and counterfactual generation.}
Deep generative models have served as structural equations for image counterfactuals over a few predefined attributes~\cite{pawlowski2020deepscm,sanchez2022diffscm,ribeiro2023pcm,pan2024cfimageediting,rasal2025semanticabduction,tong2025causaladapter}.
A recent work also brings the counterfactual view to video editing through a digital twin of the scene~\cite{shen2025cwmdt}.
However, its twin lacks explicit dependencies among objects, and without the exogenous noise of the source video each edit simply resamples a new video.
By keeping both, the Causal Diorama treats each instruction as an intervention and regenerates the video under the same exogenous noise to obtain its counterfactual.

\section{Method}
\label{sec:method}
\begin{table}[t]
 \centering
 \footnotesize
 \setlength{\tabcolsep}{6pt}\renewcommand{\arraystretch}{1.25}
 \setlength{\arrayrulewidth}{0.4pt}
 \begin{tabular}{!{\vrule width 1pt}c|c|c!{\vrule width 1pt}}
 \noalign{\hrule height 1pt}
\textbf{Causal Diorama} & \textbf{Not intervened} & \textbf{Intervened} \\   %
 \hline
 \textbf{From prompt} & Generation & Revision \\
 \hline
\textbf{From video} & \textcolor{gray}{Reconstruction} & Editing \\
 \noalign{\hrule height 1pt}
 \end{tabular}
 \caption{\textbf{Classification of the three tasks.} Gray marks the combination that we leave out.}
 \label{tab:tasks}
 \vspace{-12pt}%
\end{table}

\begin{figure*}[t]
 \centering
 \includegraphics[width=0.96\textwidth]{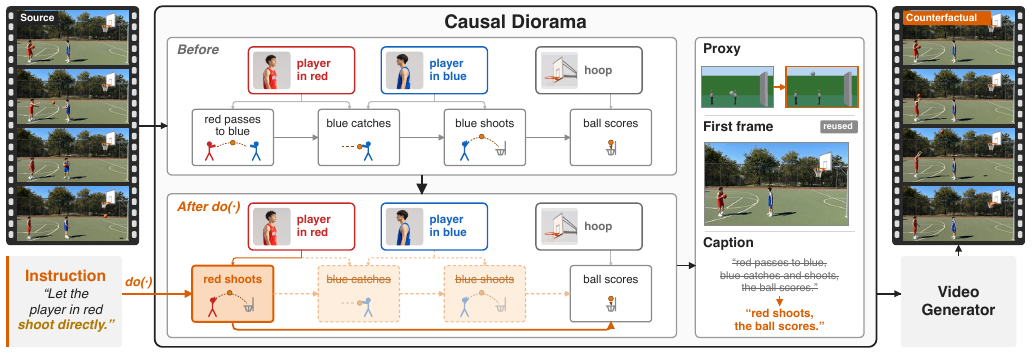}
 \vspace{-6pt}
 \caption{Overview of the intervention on the Causal Diorama, shown on a real-video edit. The instruction intervenes on an interaction node of the concept graph inferred from the source video, replacing the pass by a shot, and the change propagates to the nodes that depend on it. The connector recomputes the conditions from the intervened graph, from which the generator produces the edited video.}
 \label{fig:method}
 \vspace{-12pt}
\end{figure*}

Our framework performs generation, revision and editing through the Causal Diorama.
As summarized in \Cref{tab:tasks}, the three tasks differ in where the Causal Diorama comes from and in whether it is intervened on.
A Causal Diorama is either built from a prompt or inferred from an existing video by abduction.
Given a Causal Diorama, an instruction is carried out as an intervention that changes the specified concept variables, after which the video is regenerated, as illustrated in \Cref{fig:method}.
Generation builds a Causal Diorama from the prompt and generates the video without intervention, revision intervenes on the Causal Diorama of a generated video, and editing intervenes on the Causal Diorama inferred from an existing video.
Regenerating an existing video without intervention amounts to reconstruction, which lies outside our scope.
In the following, \Cref{sec:diorama} describes the Causal Diorama and the intervention on it, and \Cref{sec:generation,sec:revision,sec:editing} describe the three tasks.

\subsection{Causal Diorama}
\label{sec:diorama}
A Causal Diorama consists of a concept graph whose variables can be intervened on individually and a connector which turns the graph into the conditions of the video generator.
An instruction is carried out as an intervention on the Causal Diorama, and the video is regenerated as its counterfactual.
We describe the three parts in turn.

The concept graph describes a video through concept variables of six categories, namely entities, interactions, the camera, the style, the background and the text.
Entities form the core of the representation, with one entity for each object in the video such as a person or an animal.
Each entity is characterized by its type, appearance, size and trajectory, together with a reference image.
When multiple entities interact, an interaction specifies which entities take part, when they interact and what they do, such as one person handing an object to another.
The remaining four categories are global and occur exactly once per video.
The camera defines the viewpoint and its motion.
The style and the background determine the overall appearance and the setting of the video, respectively.
The text records any other constraints in natural language.
The concept graph also records the dependencies among these variables as directed edges with short reasons.
These dependencies vary across videos, so each concept graph records its own.

The connector links the concept graph to the video generator through three conditions, the proxy, the caption and the first frame, which aggregate the scattered concept variables into representations that the generator understands.
The proxy is computed by fixed code, which places a simple 3D shape for each entity and each background element, moves the entities along their trajectories and renders per-frame depth and semantic maps from the camera.
Driven by the entity trajectories and the camera, the proxy makes spatial relations explicit, which video generators often get wrong when guided by text alone~\cite{liu2026spatialalign}.
The caption is formed by filling the descriptions of the concept variables into a fixed template.
The first frame is then generated from the proxy, the caption and the reference images.
The video is finally generated from the three conditions and the reference images under the exogenous noise.
The connector is the same for every prompt and instruction and makes no decision of its own, so every property of the video is decided in the concept graph and only carried to the generator.
A more detailed explanation is provided in \Cref{sec:suppl_structure}.

Each instruction is translated into an intervention, a $\mathrm{do}$ operation on the concept graph, as illustrated in \Cref{fig:method}.
The operation assigns new values to the specified variables and propagates the change along the recorded edges to the variables that depend on them, for example by removing a deleted entity from the interactions that involve it, while all other variables retain their values.
The connector itself is left untouched and merely recomputes the three conditions from the modified graph, from which the generator regenerates the video under the same exogenous noise.
The regenerated video is the counterfactual of the original under the intervention, and differs from it only through the intervened variables and their dependents.

\subsection{Generation}
\label{sec:generation}

Given a prompt, generation first builds a Causal Diorama and then renders it into a video.
A language model agent built on Claude Opus 5.5~\cite{anthropic2026opus55} derives the concept graph from the prompt, along with the requirements that the video must satisfy, such as a dog moving from left to right.
Qwen-Image 2.1~\cite{wu2025qwenimage,qwen2026image21} then synthesizes a reference image for each entity and an image of the empty background from their descriptions, completing the concept graph.
Once the connector has rendered the proxy, a rule-based checker verifies each requirement on it, and the agent revises the concept graph until all requirements hold, so that errors are caught before any video is generated.
Finally, the video is generated by the released MiniMax H3 Ref2VA model~\cite{minimax2026h3}, equipped with the public LoRA of Code World Model~\cite{chen2026cwm} to accept the proxy as a control signal, from a fixed initial noise that we record as the exogenous noise of the video.

\subsection{Revision}
\label{sec:revision}

Revision modifies a video generated by our framework according to an instruction, so its Causal Diorama and exogenous noise are already known.
Given the instruction, the language model agent first translates it into a minimal set of changes that modify, add or remove concepts, and propagates them along the recorded dependencies to the concepts that depend on the changed ones.
Accordingly, reference images are updated for entities whose appearance changes and synthesized for new entities.
The revised graph is then checked on its proxy against the requirements that the agent adds for the revised video.
Once verified, the connector recomputes the conditions.
Since the first frame anchors the regenerated video, it is regenerated if the camera changes, edited with Qwen-Image 2.1 if only its appearance changes, and reused otherwise.
With the updated conditions, the video is finally regenerated from its original initial noise.

\subsection{Editing}
\label{sec:editing}

Editing modifies an existing video according to an instruction.
Unlike revision, the Causal Diorama and the exogenous noise of the video are unobserved and must first be recovered through abduction.
To recover the Causal Diorama, we detect and segment the entities in sampled frames with Grounding DINO~\cite{liu2023groundingdino} and SAM 2.1~\cite{ravi2024sam2}, and estimate their metric depth with Depth Anything V2~\cite{yang2024depthanythingv2}.
The camera trajectory is estimated from points tracked outside the entities by optical flow~\cite{lucas1981lk}, with their scale recovered from the metric depth.
Conditioned on these observations and the measured camera, the agent fits a Causal Diorama that reproduces the source video and records the dependencies among its variables.
To verify the fit, we render its proxy video and compare the rendered entities with the detections, and the agent refines the Causal Diorama until they agree.
The first frame and the reference images are directly extracted from the source video.

Once the Causal Diorama is recovered, abduction reduces to estimating the exogenous noise.
We obtain it by edit-friendly inversion~\cite{hubermanspiegelglas2024editfriendly} adapted to rectified flow~\cite{xie2025dnaedit}.
Conditioned on this Causal Diorama, we construct a straight path from the source latent to a fixed Gaussian noise and record the residual between this path and the generator update at each step.
Specifically, let $z$ be the latent of the source video and $c$ the conditions derived from the recovered Causal Diorama.
The generator samples with Euler steps ${z_{i+1}=z_i+\Delta\sigma_i\,v_\theta(z_i,\sigma_i,c)}$, where $\sigma_i$ decreases from $\sigma_0=1$ to $\sigma_N=0$ and $\Delta\sigma_i=\sigma_{i+1}-\sigma_i$.
Given a fixed Gaussian noise $\epsilon$, we place the sampling states on the straight path between $z$ and $\epsilon$,
\begin{equation}
 z_i=(1-\sigma_i)\,z+\sigma_i\,\epsilon,
 \label{eq:path}
\end{equation}
and record the residual of each step,
\begin{equation}
 o_i=z_{i+1}-z_i-\Delta\sigma_i\,v_\theta(z_i,\sigma_i,c).
 \label{eq:offset}
\end{equation}
Sampling from $\epsilon$ with these residuals reproduces $z$ exactly, so $\epsilon$ and $\{o_i\}$ together form the exogenous noise of the video.
After the intervention, which yields the conditions $c'$ of the revised Causal Diorama, we sample from the same noise with the same residuals,
\begin{equation}
 z'_0=\epsilon,\quad z'_{i+1}=z'_i+\Delta\sigma_i\,v_\theta(z'_i,\sigma_i,c')+o_i,
 \label{eq:edit}
\end{equation}
and decode $z'_N$ into the edited video.

Without any intervention, this process returns the source latent $z$ unchanged, which ensures that every change in the edited video is attributable to the intervention.
Moreover, the exogenous noise is determined solely by the source video and its recovered Causal Diorama, so a single inversion can serve an arbitrary number of edits.

\section{Experiments}
\label{sec:experiments}

\subsection{Benchmarks and Metrics}
\label{sec:setup}

We compare with existing methods on T2V-CompBench for generation, on CoVEBench for revision and editing, and on OpenVE-Bench for editing.
Details of the benchmarks and metrics are provided in \Cref{sec:suppl_eval}.

\textbf{T2V-CompBench}~\cite{sun2024t2vcompbench} measures controllability in seven categories covering consistent attributes, dynamic attributes, spatial relationships, motion, actions, interactions and numeracy.
Following the official protocol, we compute the scores with LLaVA-v1.6-34B~\cite{liu2024llavanext}, GroundingDINO~\cite{liu2023groundingdino}, Depth Anything~\cite{yang2024depthanything} and DOT~\cite{lemoing2024dot}.

\textbf{CoVEBench}~\cite{wu2026covebench} pairs each source video with a complex instruction that often combines multiple edits.
Revision starts from videos that our method generates from the source descriptions, while editing operates on the real source videos.
The primary metrics UAS, VQR and SEM measure instruction compliance, video quality and the preservation of unedited content, respectively.
The auxiliary metrics IFS and VRS separately measure edit execution and visual realism, both of which UAS requires.
Following the official protocol, we compute these metrics with Qwen3.5-122B-A10B and VisualQuality-R1~\cite{wu2025visualqualityr1}.

\textbf{OpenVE-Bench}~\cite{he2026openve} covers eight common edit types, from global style and background to local change, removal and addition, as well as subtitle, camera and creative edits.
Each instruction targets a single type, which allows every type to be assessed separately.
Gemini 2.5 Pro compares each result with its source video and rates instruction compliance, consistency and visual quality from 1 to 5, and the overall score combines the three ratings.

\subsection{\texorpdfstring{\texorpdfstring{Controllable Generation}{Controllable Generation}}{\texorpdfstring{Controllable Generation}{Controllable Generation}}}
\label{sec:exp_generation}

We compare our method with commercial and open-source video generators on T2V-CompBench in \Cref{tab:compbench}.
Since Wan2.2 A14B performs best among open-source generators, we also include the three strongest methods built upon it.
These methods steer attention inside the generator in a training-free manner (Comp-Attn~\cite{zhang2026compattn}), animate a first frame synthesized from a language model description (AVG~\cite{hassan2025avg}) or fine-tune the generator on captions rewritten by a learned prompt enhancer (CAPE-T2V~\cite{jia2026cape}).

\input{tables/compbench}

As our generator MiniMax H3 differs from these base models, we reimplement the training-free AVG on it, while CAPE-T2V is excluded since its weights are unavailable and reproducing it requires retraining both its prompt enhancer and generator.
We also reimplement CWM~\cite{chen2026cwm} following its paper, as our generator adopts its public LoRA.
As all other components are identical, this comparison also serves as the ablation without the Causal Diorama for generation.
For a fair comparison, all methods use Claude Opus 5.5 as the language model and Qwen-Image 2.1 as the image model.
All methods on MiniMax H3 are run by us under identical settings, while the other results are taken from T2V-CompBench and the original papers.

Our method attains the highest average score of 0.726, surpassing the best reported score of 0.667 achieved by CAPE-T2V.
It also outperforms all models with published results in five of the seven categories, with the largest margins in motion and numeracy.
On MiniMax H3, our method raises the average score of the native model from 0.500 to 0.726, well above AVG and CWM.
Methods on MiniMax H3 lag behind those on Wan2.2 A14B in action and interaction, reflecting the weaker base model in these categories.
Indeed, AVG also drops in both categories when transferred from Wan2.2 A14B to MiniMax H3.
Under the same base model, our method performs on par with AVG in action and outperforms it in interaction.
\Cref{fig:teaser}(a) shows a qualitative comparison, and more examples are provided in \Cref{fig:qual_gen}.

\subsection{Revision of Generated Videos}
\label{sec:exp_revision}

As no prior work addresses revision, we compare with four open-source instruction-based video editors, Lucy-Edit~\cite{decart2025lucyedit}, Kiwi~\cite{lin2026kiwiedit}, Ditto~\cite{bai2025ditto} and OmniWeaving~\cite{pan2026omniweaving}.
These editors directly modify the generated source video, whereas our method intervenes on its Causal Diorama and regenerates the video under the same exogenous noise.
We run all baselines with their official implementations and default settings.
As shown in \Cref{tab:covebench_rev}, our method outperforms all baselines by a large margin in UAS, IFS and VRS, and also achieves the best VQR and SEM.
Qualitative comparisons are presented in \Cref{fig:teaser}(b) and \Cref{fig:qual_rev}.

\input{tables/covebench_rev}

We further conduct two ablations on MiniMax H3 in \Cref{tab:ablation}.
First, we directly feed the source video and the instruction to MiniMax H3, which completes far fewer edits than our method.
Second, the variant without the Causal Diorama minimally edits the source description following the instruction and regenerates the video under the same noise, mirroring our revision pipeline.
It removes the whole Causal Diorama, as the connector makes no decision itself.

This variant fails to preserve much of the unedited content, which quantifies the effect in \Cref{fig:oneword}.
A text description entangles all concepts, so editing even a few words alters the entire generation.
In contrast, our method intervenes only on the concept variables specified by the instruction, so the edited video differs from the source only where the instruction requires.

\input{tables/ablation}

\subsection{Editing of Real Videos}
\label{sec:exp_editing}

Unlike revision, real-video editing requires recovering both the Causal Diorama and the exogenous noise from the input video.
Our method infers them through abduction and then intervenes as in revision.
\Cref{tab:covebench_edit} compares our method with four commercial and nine open-source video editors on CoVEBench.
We query Kling 3.0 Omni and Seedance 2.0 through their official APIs and run Bernini with its official implementation, while the results of the other editors are taken from the CoVEBench paper~\cite{wu2026covebench}.
We evaluate these three editors and our method under the official protocol.

\input{tables/covebench_edit}

Our method outperforms the open-source editors on every metric.
It reaches a UAS of 79.8 against 65.3 for Bernini, the strongest among them.
Commercial editors lag well behind our method in instruction compliance and perform comparably in video quality and video fidelity.
None of them surpasses ours in both VQR and SEM.
As CoVEBench defines no overall score, we compare all methods by their average rank over the three primary metrics, where our method ranks first.
\Cref{fig:teaser}(c) and \Cref{fig:qual_edit} further compare our method with the baselines on real videos with complex instructions.

\begin{figure}[t]
 \centering
 \includegraphics[width=\columnwidth]{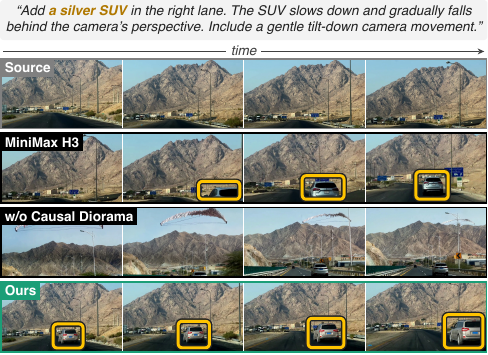}
 \caption{Qualitative ablation on CoVEBench. With MiniMax H3, the SUV moves in the direction opposite to the instruction, while the variant without the Causal Diorama changes the background. Our method follows the instruction and keeps the rest unchanged.}
 \label{fig:qual_ablation}
 \vspace{-18pt}
\end{figure}

\Cref{tab:ablation} also reports the two ablations of \Cref{sec:exp_revision} on real videos.
MiniMax H3, which directly modifies the real video as in revision, falls far short of our method in instruction compliance.
For the variant without the Causal Diorama, we recover the unknown exogenous noise with the same inversion as ours but condition it only on the source description.
We then regenerate the video from the edited description used in revision.
This variant largely fails to preserve the unedited content.
As in revision, the unedited content changes because an edit to the entangled description perturbs the whole generation process.
\Cref{tab:category} shows that this failure is consistent across all categories of edits.
Only our method follows the instruction and preserves the source video at the same time.
\Cref{fig:qual_ablation} visualizes this contrast on a real video into which an SUV is added.

\begin{figure}[t]
 \centering
 \includegraphics[width=\columnwidth]{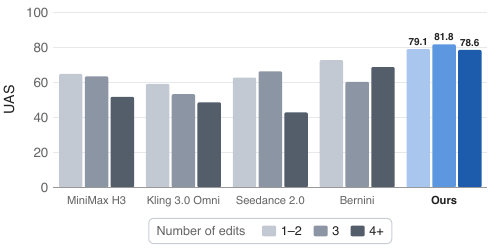}
 \caption{UAS for editing real videos on CoVEBench, grouped by the number of edits per instruction.}
 \label{fig:complexity}
 \vspace{-18pt}
\end{figure}

Moreover, the agent translates each instruction into explicit changes of individual concepts, so our method naturally scales to instructions that combine multiple edits.
To verify this, we break down UAS by the number of edits per instruction in \Cref{fig:complexity}.
All baselines perform worse on instructions with four or more edits than on those with one or two, whereas our method remains stable and extends its lead.
Notably, MiniMax H3 also degrades on complex instructions although it serves as our base model, indicating that our framework brings this robustness.

\begin{figure*}[t]
 \centering
 \includegraphics[width=\textwidth]{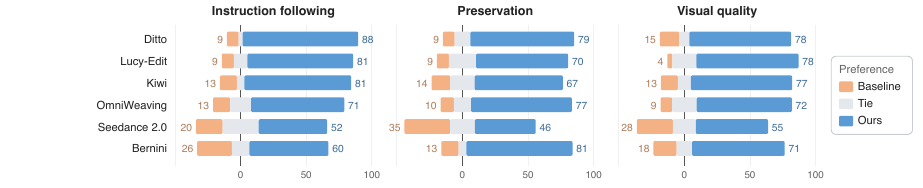}
 \caption{User study on editing real videos from CoVEBench. Each bar gives the percentages of comparisons in which participants prefer the baseline, rate a tie or prefer our method. From top to bottom, the numbers of comparisons are 67, 77, 70, 69, 69 and 68.}
 \label{fig:user_study}
 \vspace{-12pt}
\end{figure*}

We further conduct a user study on real videos from CoVEBench, collecting 420 comparisons from 42 participants.
Each participant answers 10 questions, with the baseline and the video of each question chosen at random.
Participants compare our result with the baseline and choose the preferred one or a tie in terms of instruction following, preservation of unedited content and visual quality.
As shown in \Cref{fig:user_study}, our method is preferred over every baseline on all three criteria.

\begin{figure}[t]
 \centering
 \includegraphics[width=\columnwidth]{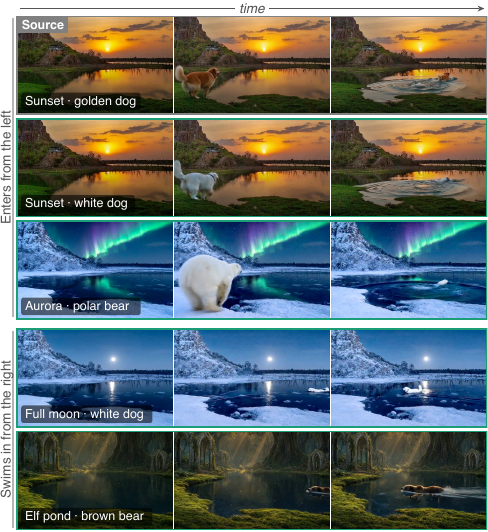}
 \caption{Edits of a real video, derived from the same Causal \mbox{Diorama} and exogenous noise. The top row shows the source video, and the remaining rows alter its scene, entity appearance and motion. Night scenes are brightened for visibility.}
 \label{fig:multiverse}
 \vspace{-18pt}
\end{figure}

Another advantage of our framework is that abduction is independent of the instruction.
The Causal Diorama and exogenous noise recovered from a video can therefore be reused for arbitrarily many edits.
Each edit may freely alter the scene, the appearance of entities and their motion, either individually or jointly.
\Cref{fig:multiverse} showcases several such combinations on a real video from CoVEBench, all derived from the same Causal Diorama and exogenous noise.

We further evaluate our method on OpenVE-Bench, a benchmark widely used to compare recent video editors~\cite{lin2026kiwiedit,pan2026omniweaving,yu2026aurora,chen2026vino,bernini2026}.
\Cref{tab:openve} compares it with three commercial and eight open-source editors using their published results.
Our method achieves the highest overall score and ranks among the top three in every category.

We also measure the inference cost of real-video editing on eight NVIDIA H20 GPUs.
Abduction, inversion and editing take 55.5 GPU seconds per megapixel-frame, about 2.3 times those of directly editing with MiniMax H3.
Excluding the agent, our editing still runs about four times as fast as the strongest open-source editor Bernini on the same GPUs.
More experiments and a detailed timing analysis are provided in \Cref{sec:suppl_results}.

\input{tables/openve}

\section{Conclusion}
\label{sec:conclusion}

We have presented Causal Dioramas, modules that consist of a concept graph whose variables can be intervened on individually and a connector that feeds the graph to a video generator.
Building on them, we developed a unified framework for generation, revision and editing.
In revision and editing, each instruction acts as an intervention that leaves the unaffected content intact.
Extensive experiments on three benchmarks demonstrate that our method surpasses both open-source and commercial models in controllability, and ablations show that the Causal Diorama is the key to preserving the unedited content.
Despite these advantages, a limitation of our method is the additional inference time of the agent, which a lighter model could reduce in future work.
We hope that, like the Diorama two centuries ago, Causal Dioramas will let video generation models control any concept of a video at will.

\section*{LLM Usage Statement}

We wrote the text and related work, designed and ran the experiments, and made the figures ourselves, using an LLM (Claude, Anthropic) to refine them.
The LLM generated parts of the code, which we reviewed, edited and approved.
We also asked it for reviewer-style feedback on the full manuscript.
No LLM was used to create data or to form research ideas, claims or conclusions.

{
    \small
    \bibliographystyle{ieeenat_fullname}
    \bibliography{main}
}

\clearpage
\maketitlesupplementary

\setcounter{section}{0}
\renewcommand{\thesection}{A\arabic{section}}
\renewcommand{\theHsection}{A\arabic{section}}
\setcounter{figure}{0}
\setcounter{table}{0}
\setcounter{equation}{0}
\renewcommand{\thefigure}{A\arabic{figure}}
\renewcommand{\thetable}{A\arabic{table}}
\renewcommand{\theequation}{A\arabic{equation}}
\renewcommand{\theHfigure}{A\arabic{figure}}
\renewcommand{\theHtable}{A\arabic{table}}
\renewcommand{\theHequation}{A\arabic{equation}}

\section{Reproducibility}
\label{sec:suppl}

\subsection{Implementation Details}
\label{sec:suppl_impl}

All language model agents are built on Claude Opus 5.5~\cite{anthropic2026opus55}.
For generation, Qwen-Image 2.1~\cite{wu2025qwenimage,qwen2026image21} synthesizes the reference images with 40 sampling steps.
Entities with the same type, label and description share one image.
The proxy video is rendered at a quarter of the output resolution and stores per-frame depth and semantics, where depth is encoded as log inverse depth clipped to 0.3 to 256~m and semantics cover 12 classes.
The video generator is MiniMax H3 Ref2VA~\cite{minimax2026h3} with the LoRA of Code World Model~\cite{chen2026cwm}.
It samples 124 frames at 24~fps and 1344\texttimes768 resolution with 20 steps and a video shift of 12.
When generating the first frame, the generator receives the empty background and up to eight entity images sorted by their area in the first frame.
When generating the full video, the first frame controls the generation together with the other conditions.

Revision reuses the generator settings of generation, so the initial noise is identical to that of the source video.
Reference images of entities whose appearance changes are edited from the original images with Qwen-Image 2.1, and the same model modifies the first frame with 40 steps.

For editing, the source video is resampled to 24~fps and center-cropped to about 1344\texttimes768 pixels with its aspect ratio preserved, and its length is set to $17n+5$ frames with at most 362 frames (15.08~s).
Perception runs on every 12th frame and the last frame, using Grounding DINO~\cite{liu2023groundingdino} with box and text thresholds of 0.25 and 0.2, SAM 2.1~\cite{ravi2024sam2} with box prompts and the metric version of Depth Anything V2~\cite{yang2024depthanythingv2}.
For the camera, up to 1,000 corners outside the entity masks are tracked by pyramidal Lucas--Kanade flow~\cite{lucas1981lk} with a 21\texttimes21 window, five pyramid levels and a forward-backward error below one pixel.
The camera poses are then solved by robust least squares and smoothed over five frames, and a camera that moves less than 0.3\textdegree{} and 3~cm is treated as fixed.
Reference images are cut out of the source frames with SAM 2.1 masks.
The inversion uses the same generator settings with 20 steps, so editing a video takes 40 generator evaluations in total.

\input{tables/category}
All experiments run on nodes with eight NVIDIA H20 GPUs (96~GB each).
Generating the first frame, generating the full video and the inversion each use all eight GPUs of a node with sequence parallelism, while Qwen-Image 2.1 and the perception models run on a single GPU.
Generating the full video peaks at about 50~GB per GPU.
For a 124-frame video, generating the full video takes about 190~s.
Each reference image takes about 24~s.
Planning a prompt takes about 70~s.
Editing a real video takes about 7~s per frame for inversion and sampling together.
Our implementation uses PyTorch 2.9.1 with CUDA 12.8 and FlashAttention-3.

\subsection{Structure of the Causal Diorama}
\label{sec:suppl_structure}

\Cref{tab:diorama_nodes} specifies the fields of the six categories of concept variables of the concept graph introduced in \Cref{sec:diorama} and of the three conditions that the connector derives from them.
For an entity, the type determines the shape of its proxy, the description specifies its appearance and the trajectory defines its position and heading over time.
An interaction specifies the participating entities, its time interval and a description of the event.
The camera is parameterized by its position, target, field of view and motion.
The background consists of a description and a reference image, together with backdrops and static props that define the layout of the empty scene.
The style is a single description, and the text describes the scene and its timeline in natural language along with the constraints not covered by other variables.
The dependencies form a list of directed edges, each with a source variable, a target variable and a short reason.
For a real video, the reference images are extracted from its frames.

\providecolor{tabNband}{RGB}{238,243,248}
\makeatletter
\@ifundefined{tabNgw}{\newlength\tabNgw\newlength\tabNuw}{}
\def\tabNmax#1#2{\settowidth\@tempdima{#2}\ifdim\@tempdima>#1\setlength#1{\@tempdima}\fi}
\def\tabNlt{0pt}\def\tabNeps{.15pt}
\@ifundefined{ver@array.sty}{\def\tabNL{\def\tabNlt{.5\arrayrulewidth}}}{\let\tabNL\relax}
\def\tabNtile#1#2{\hspace*{\dimexpr\tabNlt-\tabcolsep\relax}%
 \rlap{\smash{\color{tabNband}\vrule width\dimexpr#1+2\tabcolsep-\tabNlt+\tabNeps\relax
 height\ht\@arstrutbox depth\dimexpr\dp\@arstrutbox+\tabNeps\relax}}%
 \hspace*{\dimexpr\tabcolsep-\tabNlt\relax}\makebox[#1][l]{#2}}
\makeatother
\begin{table*}[t]
 \centering
 \small
 \setlength{\tabcolsep}{4pt}\renewcommand{\arraystretch}{1.16}
 \setlength{\aboverulesep}{0pt}\setlength{\belowrulesep}{0pt}
 \def\tabNcw{0.26\textwidth}\def\tabNfw{0.49\textwidth}%
 \setlength\tabNgw{0pt}\tabNmax\tabNgw{Dependencies}\tabNmax\tabNgw{Variable}\tabNmax\tabNgw{Interactions}\tabNmax\tabNgw{Background}\tabNmax\tabNgw{Proxy}\tabNmax\tabNgw{First frame}%
 \setlength\tabNuw{0pt}\tabNmax\tabNuw{Unique}%
 \begin{tabular}{l|p{\tabNcw}|p{\tabNfw}|c}
 \toprule
 \makebox[\tabNgw][l]{Item} & \tabNL Content & Main fields & \tabNL Unique \\
 \midrule
 \tabNtile{\tabNgw}{\textit{Concept graph}} & \tabNL\tabNtile{\tabNcw}{} & \tabNtile{\tabNfw}{} & \tabNL\tabNtile{\tabNuw}{} \\
 \makebox[\tabNgw][l]{Entities} & \tabNL Objects in the scene & type, description, size, trajectory, reference image & \tabNL No \\
 \makebox[\tabNgw][l]{Interactions} & \tabNL Events between entities & entities, time interval, description & \tabNL No \\
 \makebox[\tabNgw][l]{Camera} & \tabNL Viewpoint and its motion & position, target, field of view, motion mode, description & \tabNL Yes \\
 \makebox[\tabNgw][l]{Background} & \tabNL Background of the video & backdrops, static props, description, reference image & \tabNL Yes \\
 \makebox[\tabNgw][l]{Style} & \tabNL Style of the video & description & \tabNL Yes \\
 \makebox[\tabNgw][l]{Text} & \tabNL Language description of the video & scene, timeline, constraints & \tabNL Yes \\
 \makebox[\tabNgw][l]{Dependencies} & \tabNL Relations among the variables & source variable, target variable, reason & \tabNL No \\
 \midrule %
 \tabNtile{\tabNgw}{\textit{Connector}} & \tabNL\tabNtile{\tabNcw}{} & \tabNtile{\tabNfw}{} & \tabNL\tabNtile{\tabNuw}{} \\
 \makebox[\tabNgw][l]{Proxy} & \tabNL Geometry of the scene & depth map, semantic map & \tabNL Yes \\
 \makebox[\tabNgw][l]{Caption} & \tabNL Text condition of a generator & descriptions filled into a template & \tabNL Yes \\
 \makebox[\tabNgw][l]{First frame} & \tabNL Starting frame of the video & image & \tabNL Yes \\
 \bottomrule
 \end{tabular}
 \caption{Structure of a Causal Diorama, the variables and dependencies of the concept graph and the conditions of the connector. Unique marks the items of which a Causal Diorama has exactly one.}
 \label{tab:diorama_nodes}
\end{table*}

\subsection{Benchmarks and Metrics}
\label{sec:suppl_eval}

\textbf{T2V-CompBench}~\cite{sun2024t2vcompbench} contains seven categories, consistent attribute binding (Cons.), dynamic attribute binding (Dyn.), spatial relationships (Spat.), motion binding (Mot.), action binding (Act.), object interactions (Inter.)\ and generative numeracy (Num.).
We run the official evaluation scripts with their default settings, and each category is scored from 0 to 1.
LLaVA-v1.6-34B~\cite{liu2024llavanext} scores consistent attribute binding, dynamic attribute binding, action binding and object interactions from the video frames.
For spatial relationships, GroundingDINO~\cite{liu2023groundingdino} detects the objects and Depth Anything~\cite{yang2024depthanything} estimates their depth to verify the described relation.
For numeracy, GroundingDINO counts the objects and compares the count with the number in the prompt.
For motion binding, GroundingDINO and SAM~\cite{kirillov2023sam} segment the objects, and DOT~\cite{lemoing2024dot} tracks them to check their direction of motion.
Avg.\ is the mean over the seven categories.

\textbf{CoVEBench}~\cite{wu2026covebench} evaluates instruction compliance, video quality and video fidelity with five metrics, of which UAS, VQR and SEM are the primary metrics of the three dimensions.
Its judge, Qwen3.5-122B-A10B, answers a checklist of questions generated for each instruction.
\begin{itemize}[leftmargin=*,itemsep=2pt,topsep=2pt]
  \item \textbf{Instruction Following (IFS)} measures whether each edit is executed, regardless of its quality. It is the accuracy of the judge on the execution questions of the checklist.
  \item \textbf{Video Realism (VRS)} measures whether the edits look natural. It is the accuracy on the realism questions of the checklist.
  \item \textbf{Union Accuracy (UAS)} measures whether each edit is both executed and natural. It counts an edit as successful only if its execution and realism questions are both answered correctly.
  \item \textbf{Comprehensive Quality (VQR)} measures the visual quality of the whole video. It is rated by VisualQuality-R1~\cite{wu2025visualqualityr1} as a holistic assessment of the video.
  \item \textbf{Semantic Consistency (SEM)} measures the preservation of the content that the instruction does not mention. The judge rates it from 1 to 10, which is converted to a score out of 100.
\end{itemize}

\textbf{OpenVE-Bench}~\cite{he2026openve} contains eight categories, Global Style (Style), Background Change (Bkg.), Local Change (Chg.), Local Remove (Rem.), Local Add (Add), Subtitles Edit (Sub.), Camera Multi-Shot Edit (Cam.)\ and Creative Edit (Cre.).
For each result, Gemini 2.5 Pro receives the source video, the edited video and the instruction, and it rates instruction compliance, consistency and detail fidelity, and visual quality and stability from 1 to 5.
The evaluation prompts are designed for each category, and the overall score of a result combines the three ratings.

\subsection{\texorpdfstring{Sources of the Reported Results}{Sources of the Reported Results}}
\label{sec:suppl_sources}
For the baselines that we do not run ourselves, we report the numbers published for them.
We take them from the paper of the respective benchmark whenever it reports the method, and otherwise from the paper of the method itself or of a method built on it, so commercial generators and editors appear in the versions evaluated there.
All methods on MiniMax H3, as well as Kling 3.0 Omni, Seedance 2.0 and Bernini on CoVEBench, are run by us under the settings of \Cref{sec:suppl_impl}.
For AVG on Wan2.2 A14B in \Cref{tab:compbench}, we report its variant without fine-tuning, which anchors generation on a first frame synthesized from text, as this is the setting of our reimplementation on MiniMax H3.
On OpenVE-Bench, the results of Kling O1 and PixVerse R1 are taken from Mamoda~2.5~\cite{shi2026mamoda25}, which does not state its judge.
In \Cref{tab:compbench}, the results of Wan2.2 A14B, LTX 2.3 and Veo 3.1 are taken from Comp-Attn~\cite{zhang2026compattn}, CAPE-T2V~\cite{jia2026cape} and VQQA~\cite{song2026vqqa}, respectively, which evaluate them as their base models.

\section{Additional Results}
\label{sec:suppl_results}

\input{tables/threshold}
\textbf{Effect of the Causal Diorama.} \Cref{tab:ablation} in the main paper compares our method with the variant without the Causal Diorama, which regenerates the video with MiniMax H3 from the source description edited according to the instruction under the same exogenous noise.
Since the requested changes are easy to realize for a strong base model such as MiniMax H3, this variant already achieves a high UAS.
Even so, our method improves UAS by 1.8 points in editing and 2.1 points in revision.
More importantly, our method improves SEM by 38.0 points in editing and 20.3 points in revision, showing a substantial advantage in preserving the unedited content.
We further count a result as a success only when its UAS and SEM both reach given thresholds, and our method achieves the highest success rate under all ten thresholds in \Cref{tab:threshold}.
Although the variant without the Causal Diorama is close to ours in mean UAS, its success rate drops much faster as the SEM threshold tightens, reaching only 8.5\% against our 47.6\% under UAS $=$ 100 and SEM $\geq$ 90.
Together, these results confirm that the Causal Diorama allows the change of each concept to be controlled independently.

\input{tables/ablation_causal}
\textbf{Causal ablations.} We further decompose our method in \Cref{tab:causal_ablation} to examine the roles of the causal mechanism and the exogenous noise.
For the variant without dependencies, we restrict the language model agent to changing only the concepts named in the instruction, without propagating the change to the concepts that depend on them.
This variant drops to 68.9 in UAS, even below the 78.0 of the variant without the Causal Diorama in \Cref{tab:ablation}.
The latter leaves the generation unconstrained, so the base model can freely adapt the rest of the video to the edited description.
In contrast, keeping the dependent concepts unchanged breaks the causal structure and leads to implausible combinations, such as an interaction that still involves a removed entity.
Moreover, the concepts that retain their original values pull the video toward the source video, which makes the requested change harder to complete.
This is also reflected by its SEM of 93.6, which stays on par with ours despite the much lower UAS.
Without inversion, the video is regenerated under newly sampled noise instead of the recovered exogenous noise, which lowers UAS by 6.6 points and SEM by 4.6 points.

\textbf{Inference cost.} We measure the inference cost of real-video editing on CoVEBench, where both our method and Bernini run on eight NVIDIA H20 GPUs.
Since the two methods produce videos of different lengths and resolutions, we report the median time per megapixel-frame.
Excluding the language model agent, our editing takes 8.5\,s, whereas Bernini takes 34.8\,s.
In terms of GPU time, our abduction, inversion and editing cost 55.5 GPU seconds, about one fifth of the 278.6 GPU seconds of Bernini.
Directly editing with MiniMax H3 on a single GPU takes 23.8 GPU seconds.
This gap mainly comes from the base model, as MiniMax H3 samples 20 steps without classifier-free guidance, whereas Bernini samples 40 steps with four-way guidance.
Even with the agent, whose wall-clock time is an upper bound that includes network latency, our editing takes 16.5\,s and remains about twice as fast as Bernini.

\textbf{Qualitative comparisons.} We provide additional qualitative comparisons on the three tasks, complementing \Cref{fig:teaser}.
For generation, \Cref{fig:qual_gen} shows that only our method turns the leaf from green to yellow, whereas the baselines keep its color unchanged.
In the second example, our method makes the monkey juggle the bananas and the teddy bear look surprised, whereas the baselines fail to show the juggling or omit the teddy bear.
For revision in \Cref{fig:qual_rev}, our method makes the man stand up and raise a glowing cane, whereas the baselines leave him seated.
In the second example, our method removes the clapping people and keeps the rest of the video intact, but most baselines leave them in the background.
For editing, \Cref{fig:qual_edit} illustrates that our method lays the woman back onto the pillows, whereas the baselines keep her sitting up.
In the second example, our method places a single black cat by the front wheel without disturbing the scene, while the baselines misplace the cat or alter the truck and the framing.

\newcommand{\qctag}[2]{{\setlength{\fboxsep}{1.5pt}\colorbox{#1}{\fontsize{6.5}{6.5}\selectfont\sffamily\bfseries\color{white}\vphantom{Ag}#2}}}
\newcommand{\qcimg}[3][black]{%
  \sbox0{\setlength{\fboxsep}{0pt}\setlength{\fboxrule}{0.8pt}\fcolorbox{#1}{white}{\includegraphics[width=\dimexpr\linewidth-1.6pt\relax]{#3}}}%
  \sbox2{\qctag{#1}{#2}}%
  \usebox0\llap{\makebox[\wd0][l]{\hspace{0.8pt}\raisebox{\dimexpr\ht0-0.8pt-\ht2\relax}{\usebox2}}}\\[1.5pt]}
\newif\ifqcdo
\newcommand{\qckey}[1]{\textcolor{figkey}{\textbf{#1}}}
\newcommand{\qcbox}[1]{{\ifqcdo\colorlet{qcacc}{figorange}\else\colorlet{qcacc}{gray}\fi
  \sbox0{\colorbox{figgray}{\parbox[c][24pt][c]{\dimexpr\linewidth-2\fboxsep\relax}{\centering\footnotesize\itshape #1}}}%
  \usebox0\llap{\makebox[\wd0][l]{\raisebox{-\dp0}{\color{qcacc}\rule{1.4pt}{\dimexpr\ht0+\dp0\relax}}}}}}

\begin{figure*}[t]
 \centering
 \begin{minipage}[t]{0.493\textwidth}\centering
  \qcbox{``The leaf is turning from \qckey{green to yellow}.\textquotedblright}\\[3pt]
  \qcimg[figgreen]{Ours}{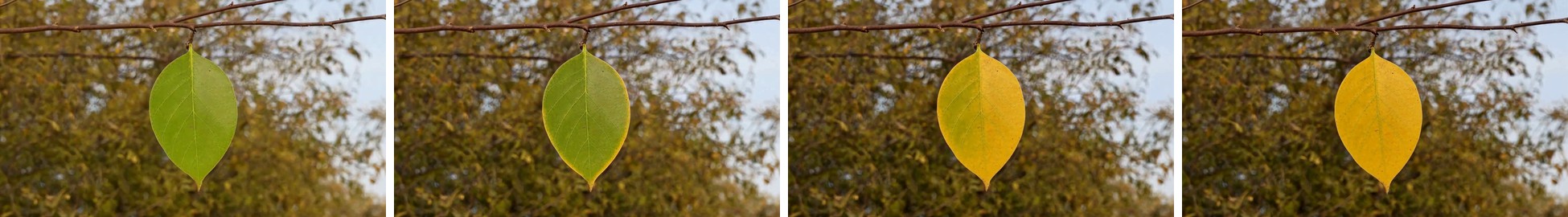}
  \qcimg{PixVerse V3}{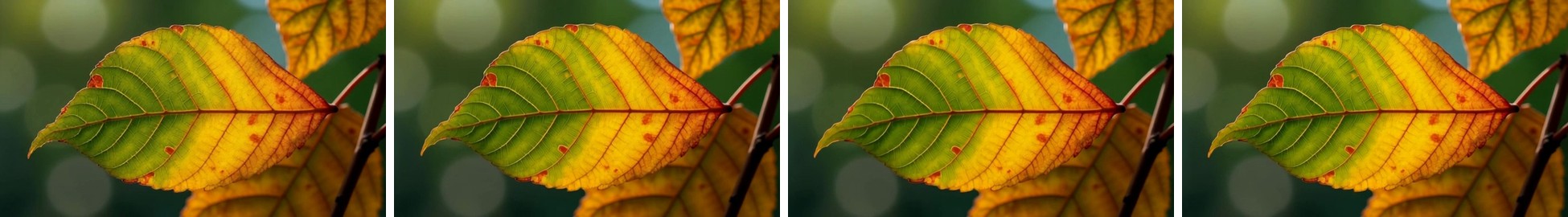}
  \qcimg{Kling 1.0}{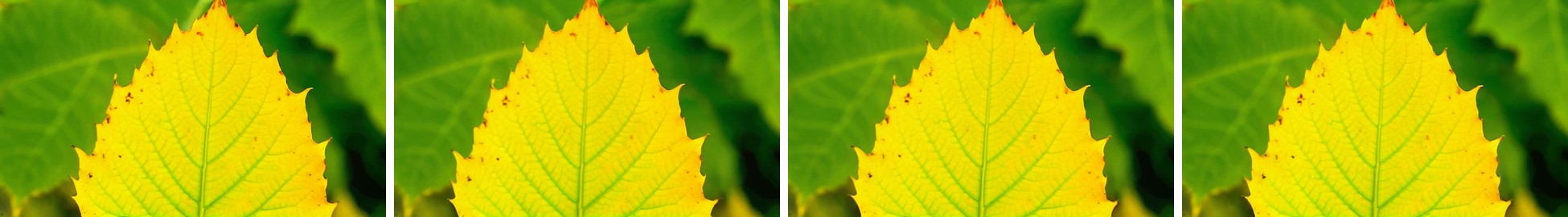}
  \qcimg{Dreamina 1.2}{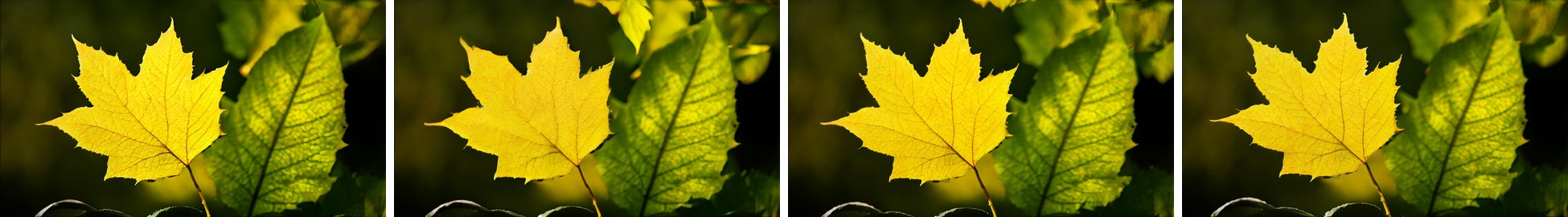}
  \qcimg{CogVideoX 5B}{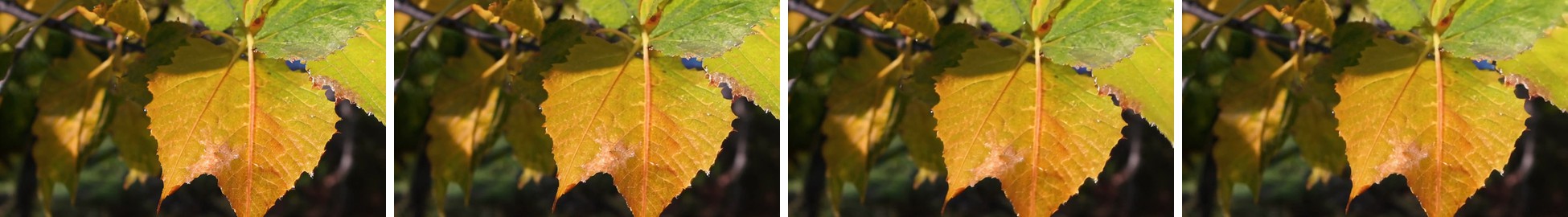}
 \end{minipage}\hfill
 \begin{minipage}[t]{0.493\textwidth}\centering
  \qcbox{``A monkey \qckey{juggles} bananas, a teddy bear is \qckey{surprised}\textquotedblright}\\[3pt]
  \qcimg[figgreen]{Ours}{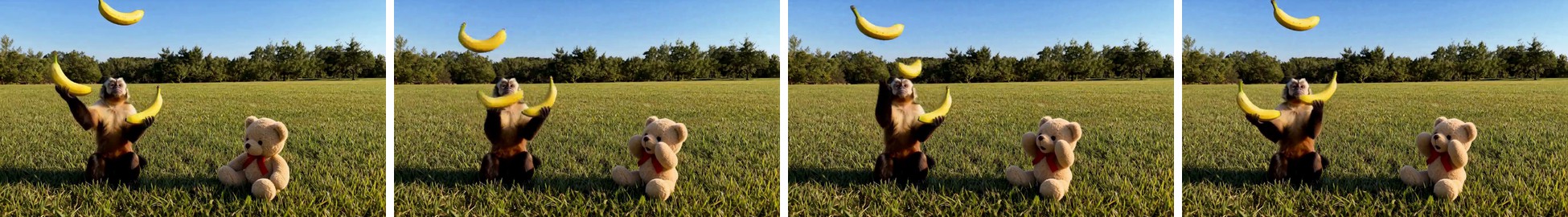}
  \qcimg{PixVerse V3}{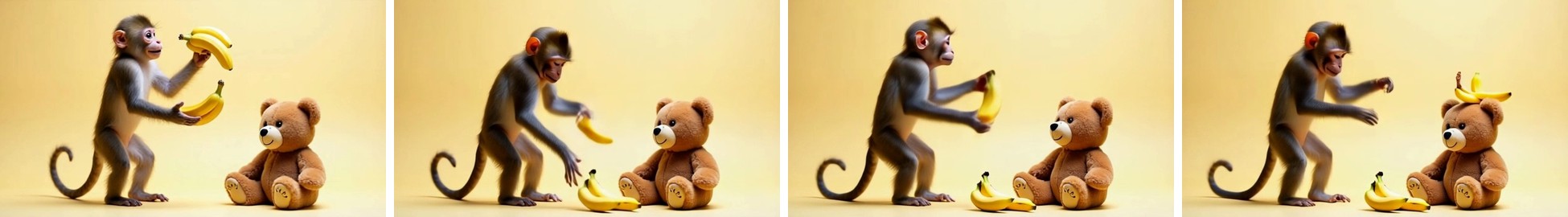}
  \qcimg{Kling 1.0}{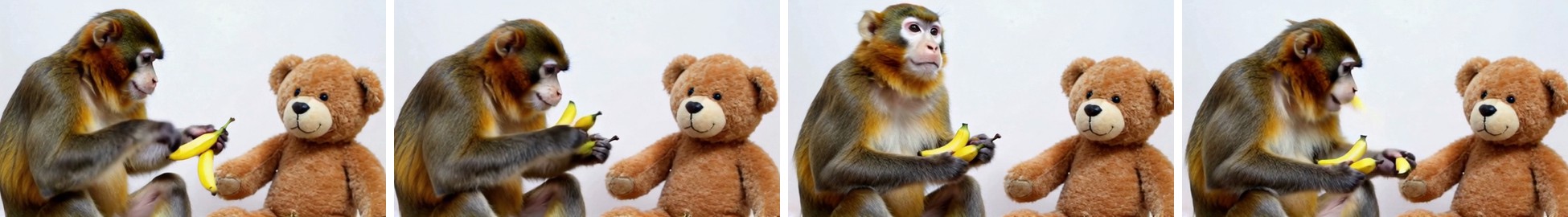}
  \qcimg{Dreamina 1.2}{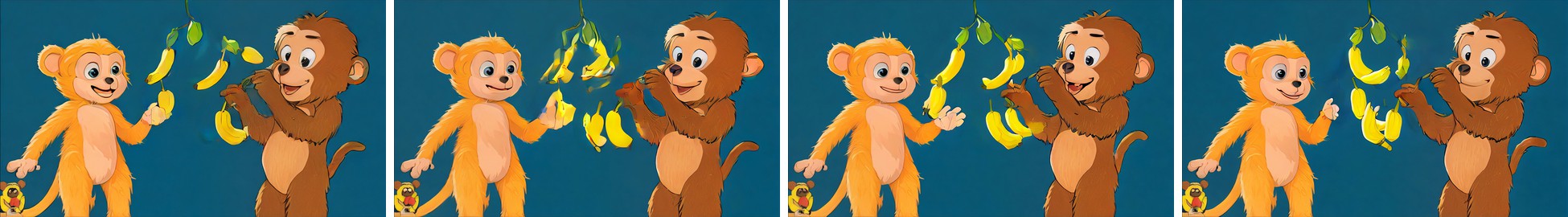}
  \qcimg{CogVideoX 5B}{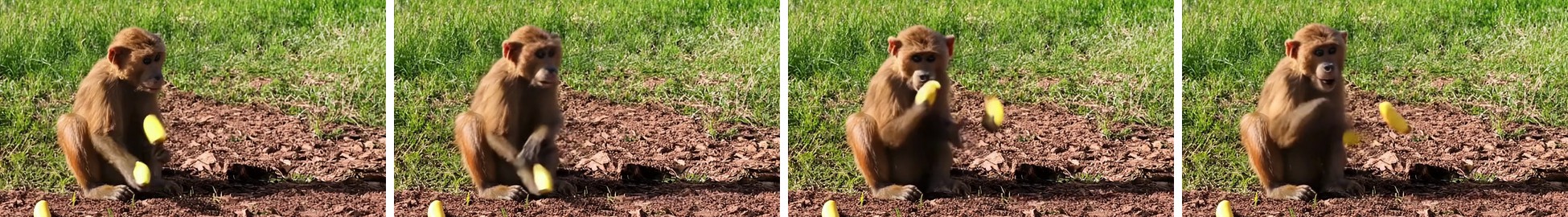}
 \end{minipage}
 \caption{Qualitative comparison of generation on T2V-CompBench.}
 \label{fig:qual_gen}
\end{figure*}

\begin{figure*}[t]
 \centering
 \begin{minipage}[t]{0.493\textwidth}\centering\qcdotrue
  \qcbox{``\qckey{Stand up} and \qckey{raise} the wooden cane high toward the sky,\\with a bright \qckey{golden glow} at its tip.\textquotedblright}\\[3pt]
  \qcimg[gray]{Source}{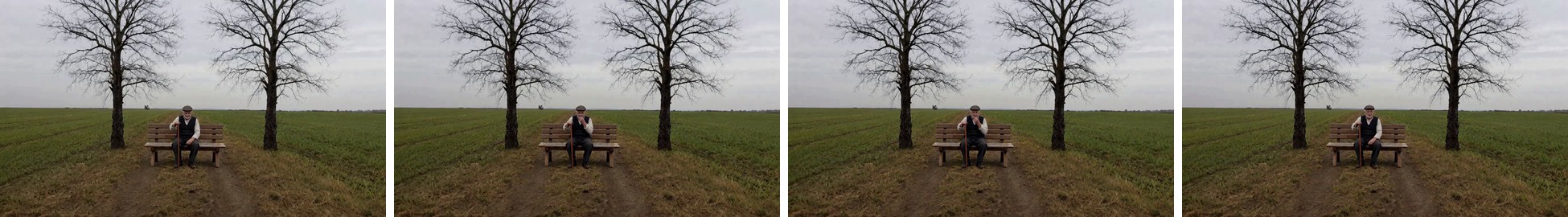}
  \qcimg[figgreen]{Ours}{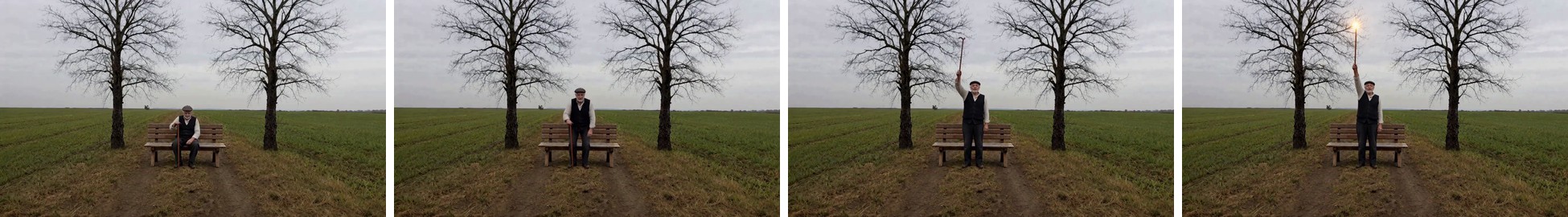}
  \qcimg{OmniWeaving}{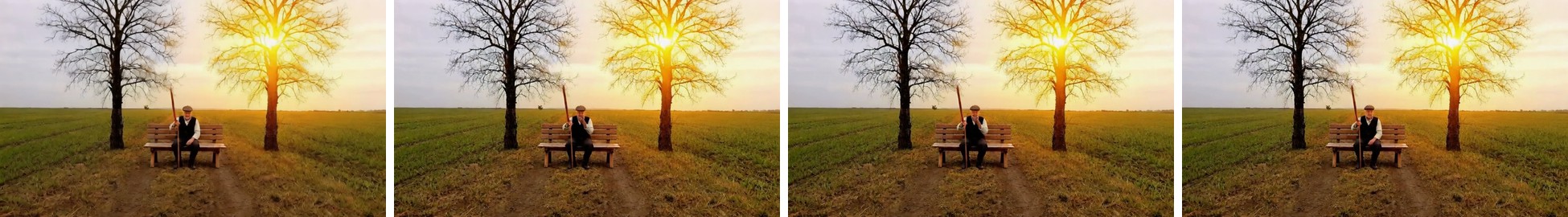}
  \qcimg{Ditto}{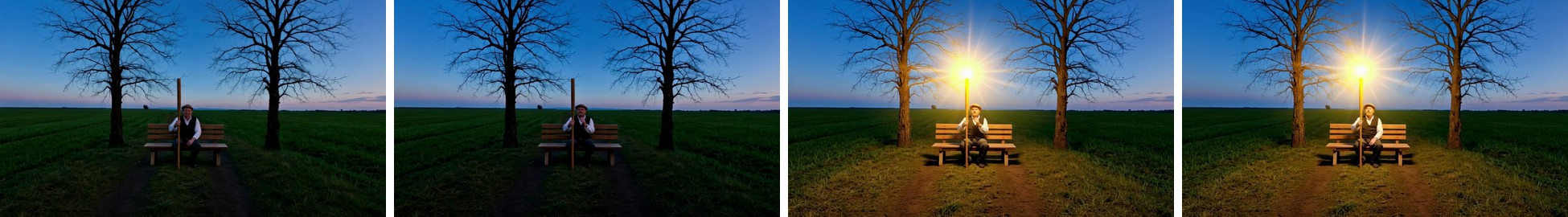}
  \qcimg{Kiwi}{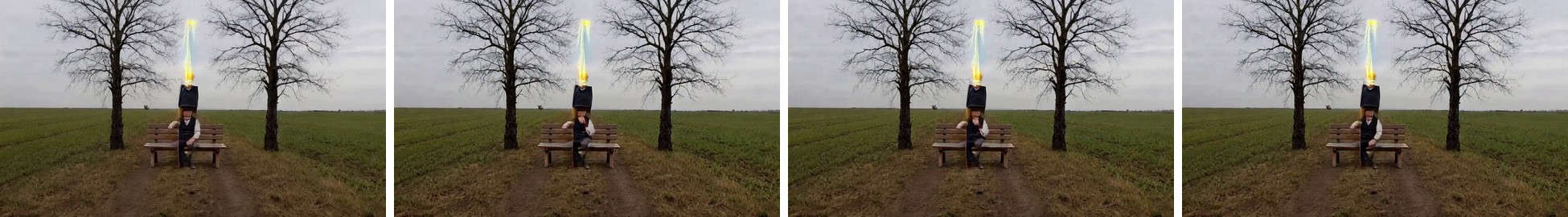}
  \qcimg{Lucy}{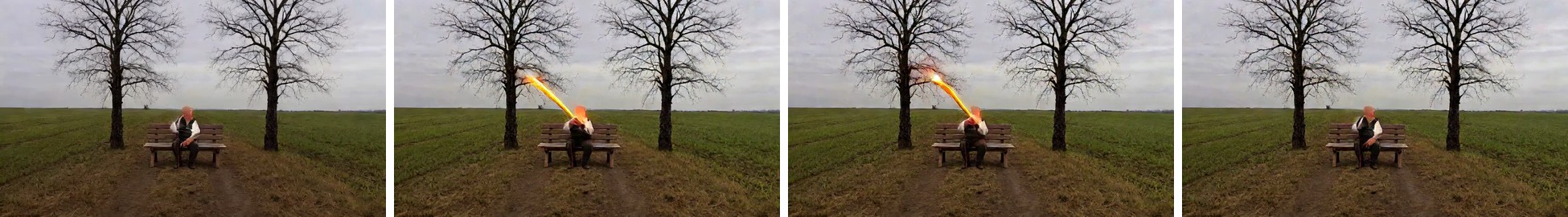}
 \end{minipage}\hfill
 \begin{minipage}[t]{0.493\textwidth}\centering\qcdotrue
  \qcbox{``\qckey{Remove the clapping people} from the background,\\keeping the man blowing out the candles.\textquotedblright}\\[3pt]
  \qcimg[gray]{Source}{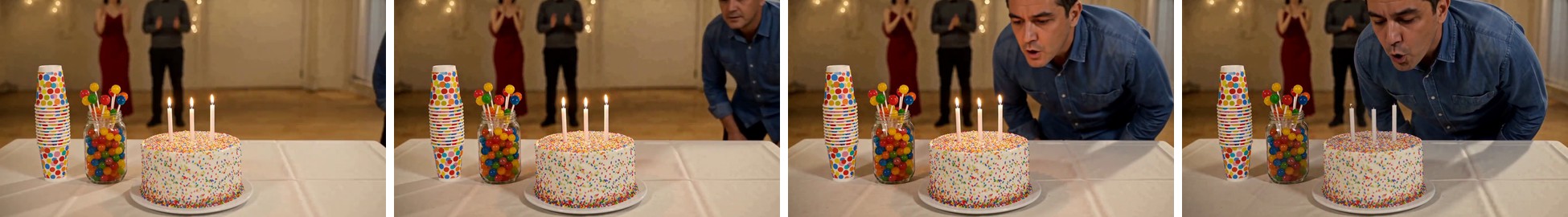}
  \qcimg[figgreen]{Ours}{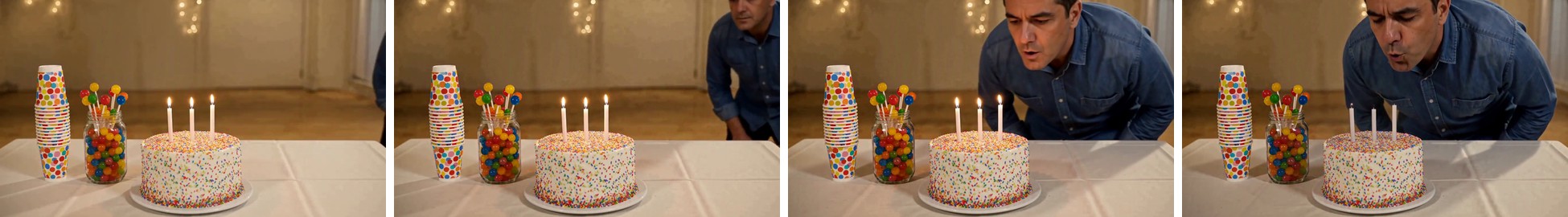}
  \qcimg{OmniWeaving}{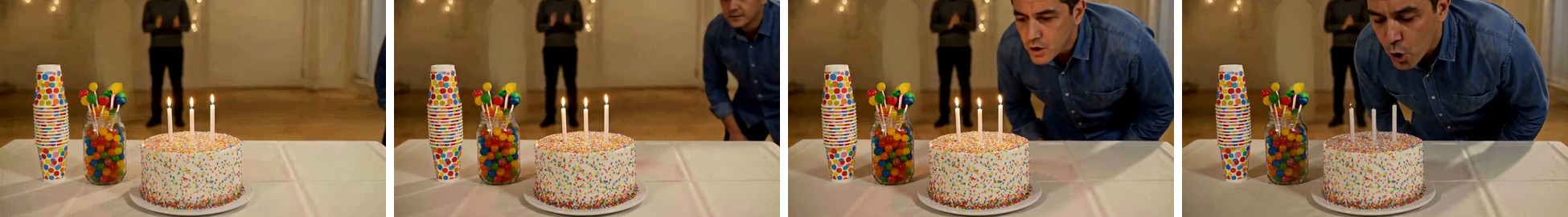}
  \qcimg{Ditto}{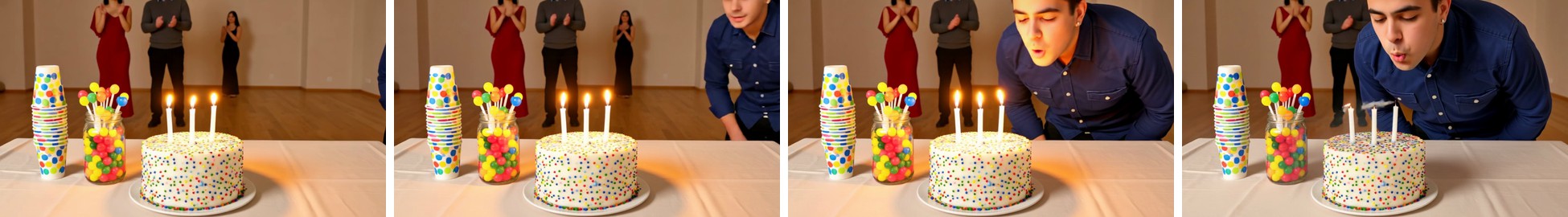}
  \qcimg{Kiwi}{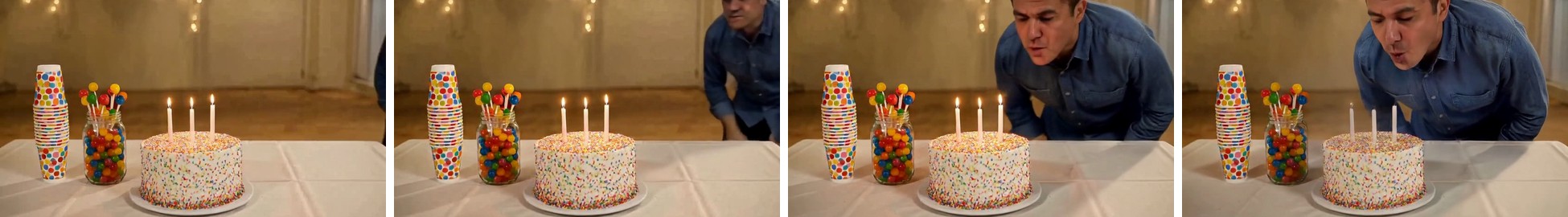}
  \qcimg{Lucy}{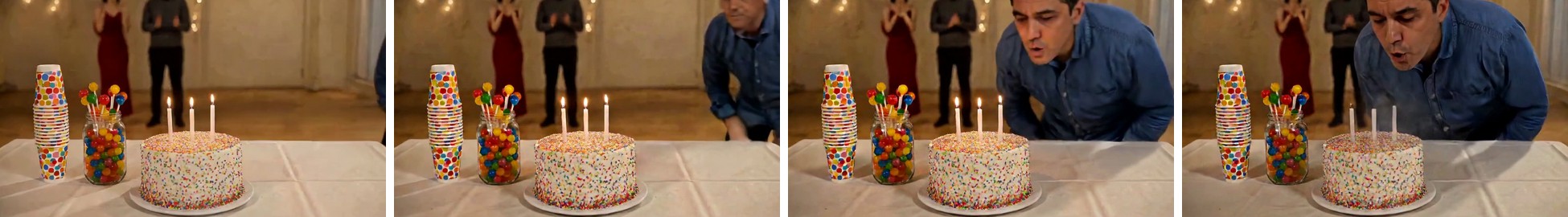}
 \end{minipage}
 \caption{Qualitative comparison of revision on CoVEBench.}
 \label{fig:qual_rev}
\end{figure*}

\begin{figure*}[t]
 \centering
 \begin{minipage}[t]{0.493\textwidth}\centering\qcdotrue
  \qcbox{``Change the action from touching her face to \qckey{placing}\\\qckey{the phone down} and \qckey{lying back} onto the pillows.\textquotedblright}\\[3pt]
  \qcimg[gray]{Source}{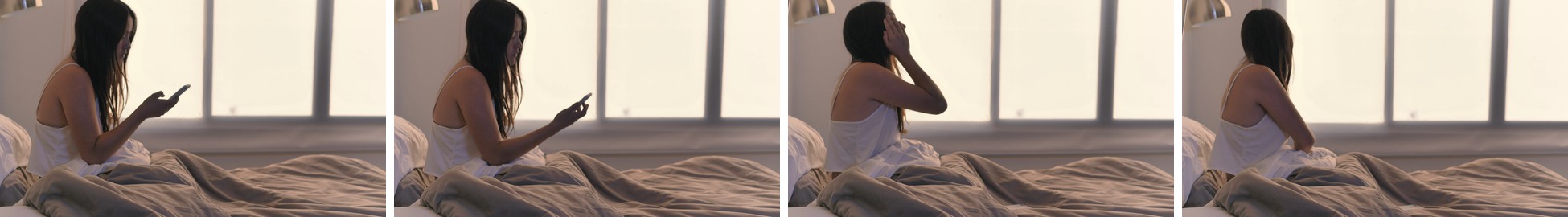}
  \qcimg[figgreen]{Ours}{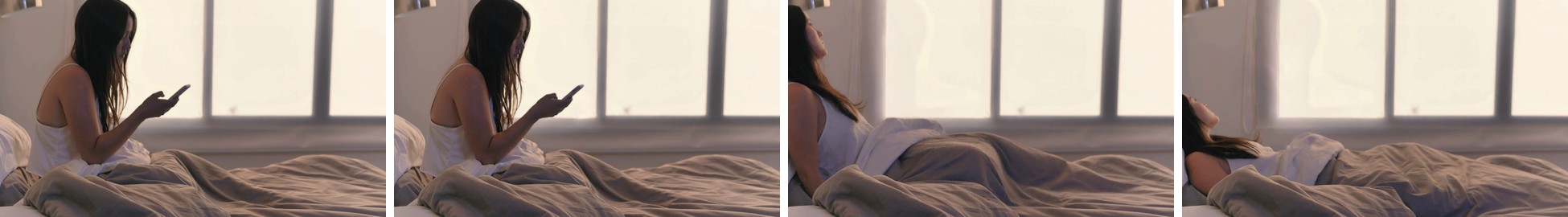}
  \qcimg{Seedance 2.0}{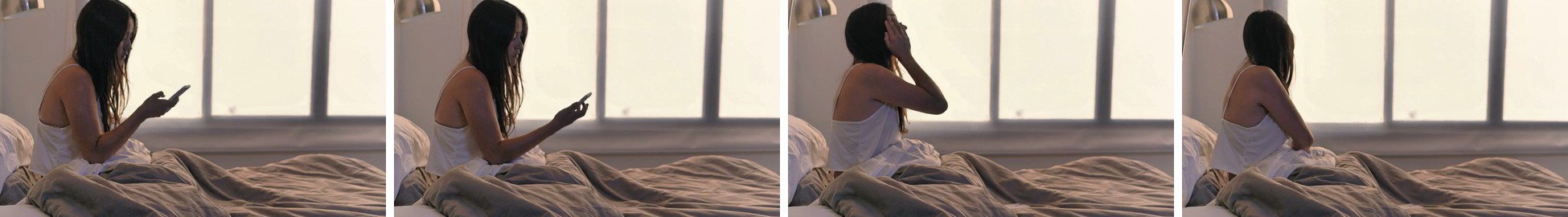}
  \qcimg{Kling 3.0 Omni}{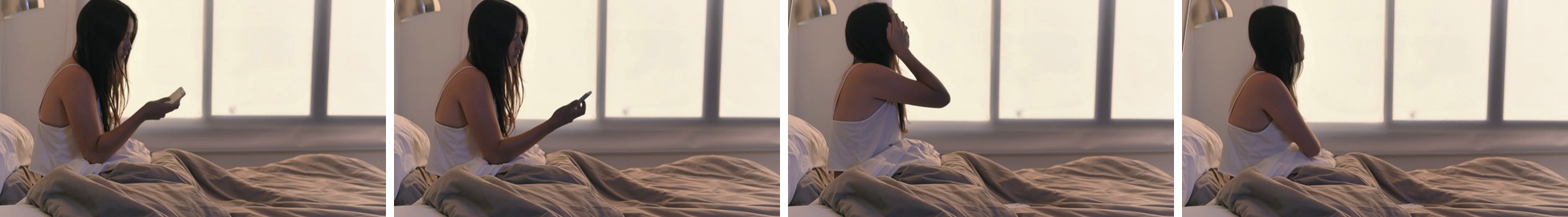}
  \qcimg{OmniWeaving}{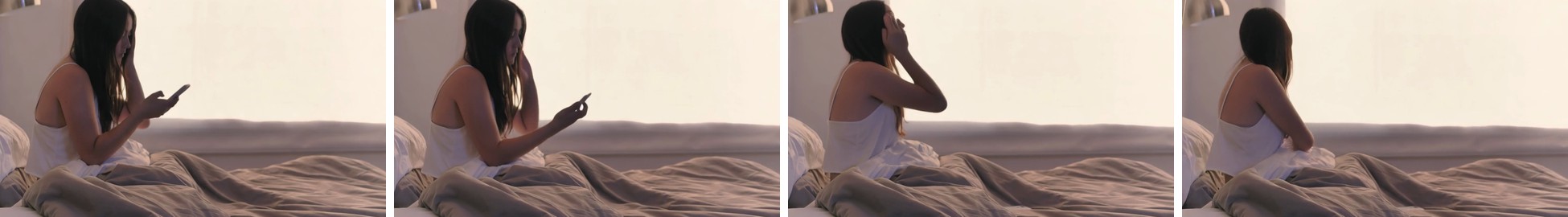}
  \qcimg{Kiwi}{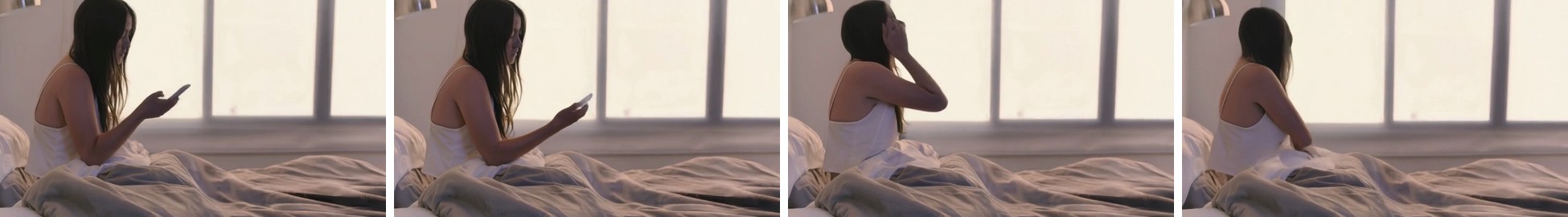}
 \end{minipage}\hfill
 \begin{minipage}[t]{0.493\textwidth}\centering\qcdotrue
  \qcbox{``Replace the cats with a \qckey{single black cat} near the front wheel,\\make the morning \qckey{foggy} and \qckey{pan slowly to the right}.\textquotedblright}\\[3pt]
  \qcimg[gray]{Source}{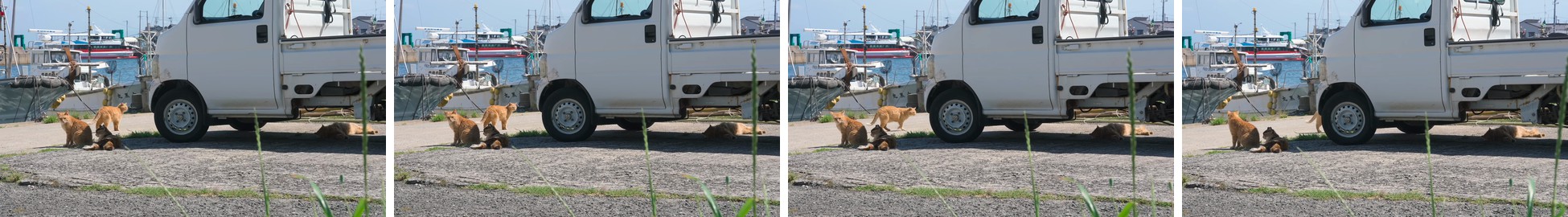}
  \qcimg[figgreen]{Ours}{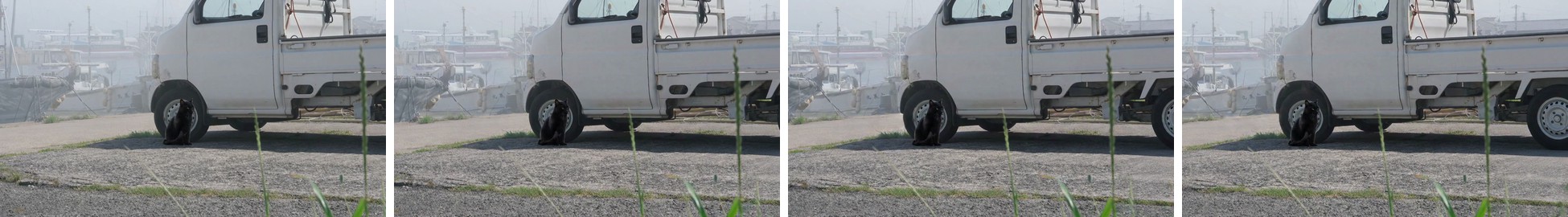}
  \qcimg{Seedance 2.0}{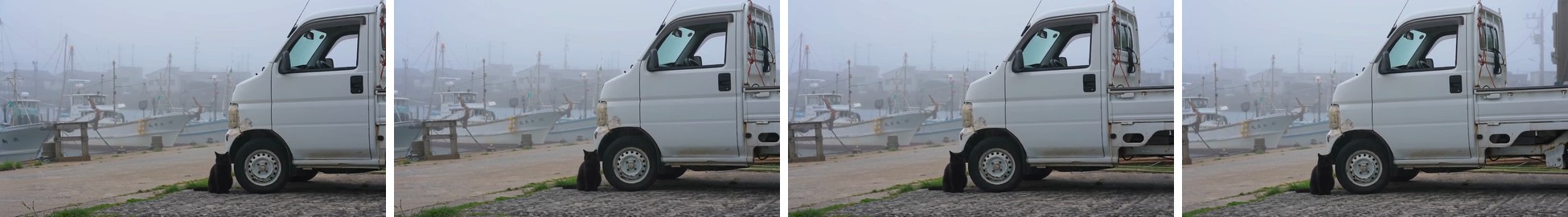}
  \qcimg{Kling 3.0 Omni}{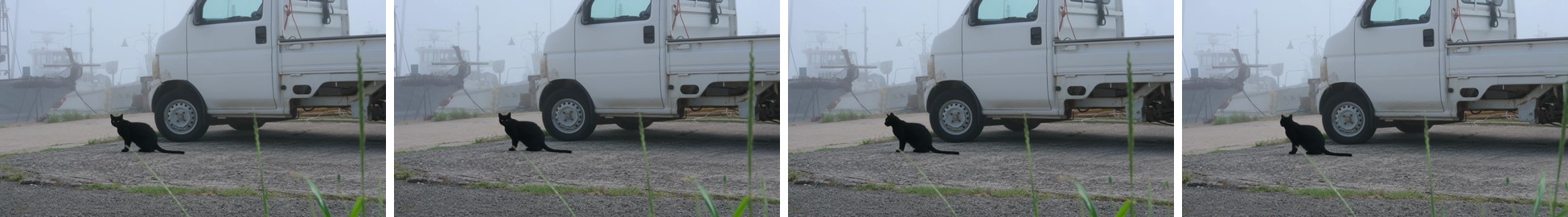}
  \qcimg{Bernini}{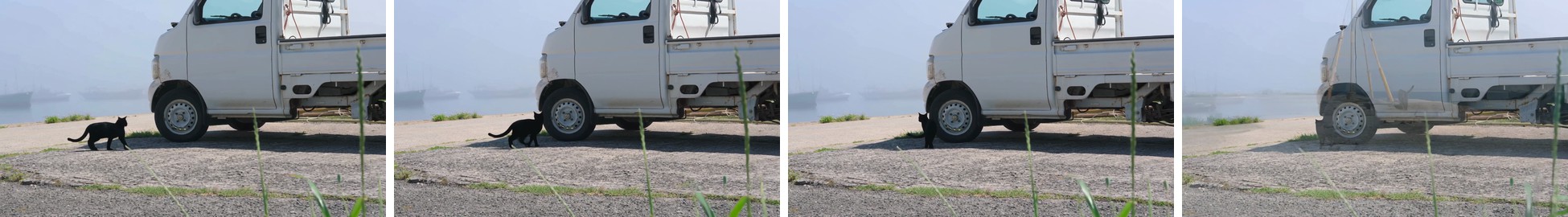}
  \qcimg{OmniWeaving}{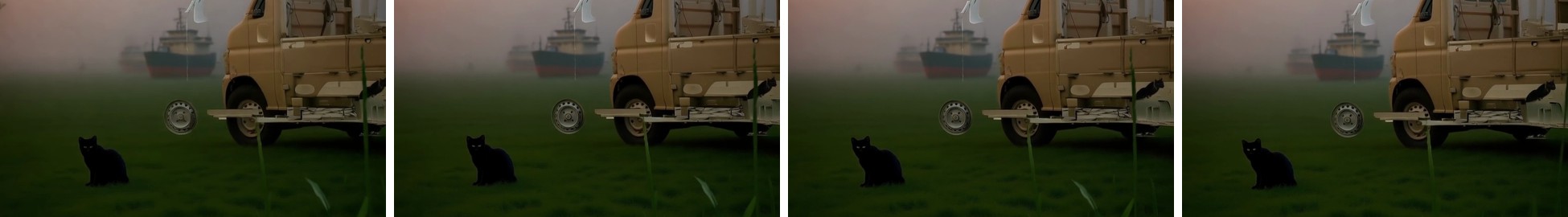}
 \end{minipage}
 \caption{Qualitative comparison of editing on CoVEBench.}
 \label{fig:qual_edit}
\end{figure*}

\section{Comparison on Shared Videos}
\label{sec:suppl_common}
\input{tables/covebench_common}
Kling 3.0 Omni and Seedance 2.0 cannot process all videos of CoVEBench, so \Cref{tab:covebench_edit} evaluates each method on its own set of videos.
\Cref{tab:common} compares each of them with our method on the videos that both process.
As in \Cref{tab:covebench_edit}, our method leads Kling 3.0 Omni and Seedance 2.0 by 26.8 and 19.6 points in UAS, respectively, and performs comparably in video quality and video fidelity.

\section{User Study}
\label{sec:suppl_userstudy}
\begin{figure*}[p]
 \centering
 \includegraphics[width=\textwidth]{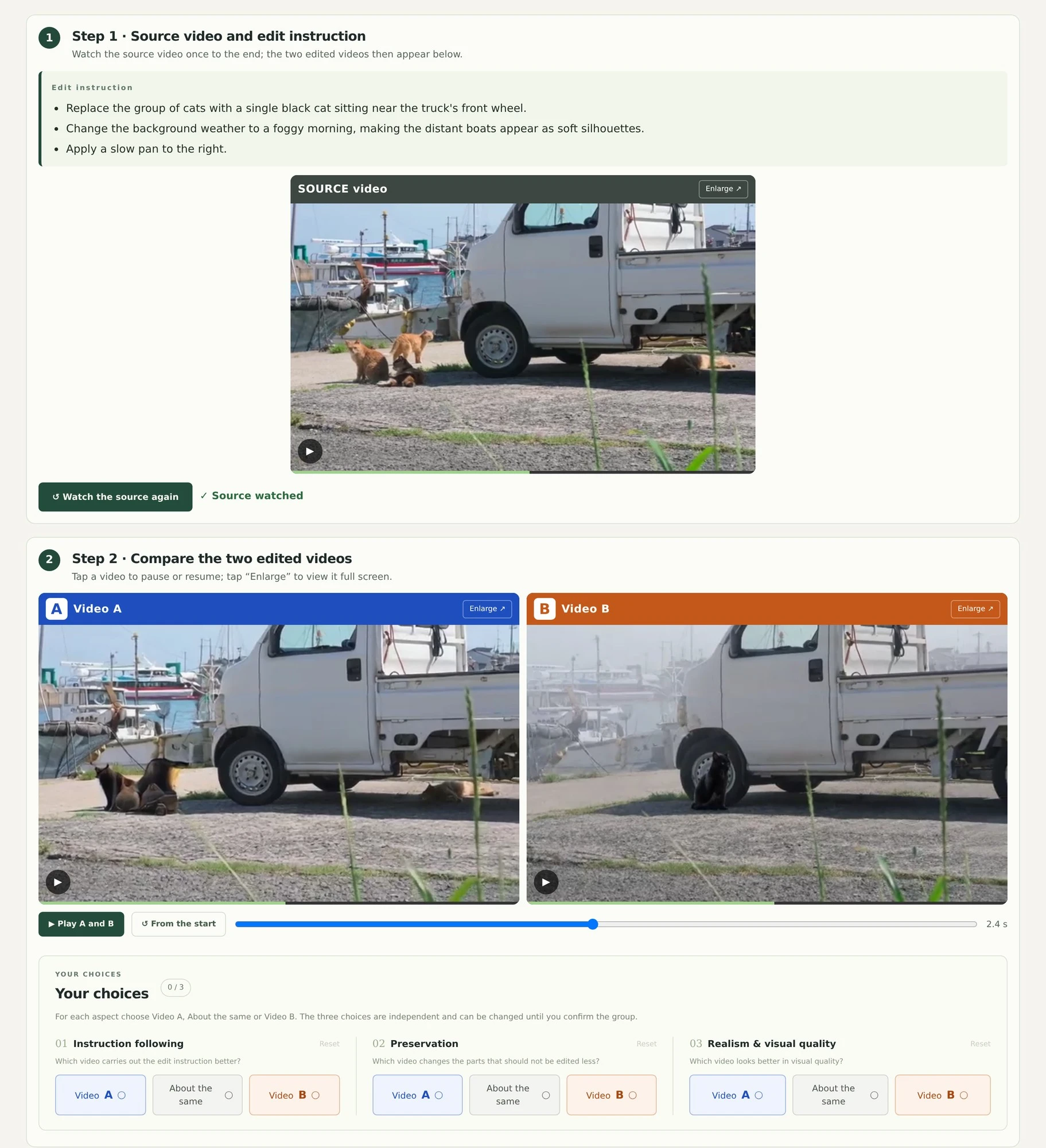}
 \caption{Interface of the user study, shown for one question. Top: the edit instruction and the source video, which the participant watches to the end before the edited videos appear. Bottom: the two edited videos, played side by side as Video A and Video B, and the three questions, each answered with Video A, About the same or Video B.}
 \label{fig:user_study_interface}
\end{figure*}

The user study in \Cref{sec:exp_editing} compares our method with six baselines, Ditto, Lucy-Edit, Kiwi, OmniWeaving, Seedance 2.0 and Bernini, on editing real videos from CoVEBench.
Each of the 42 participants answers 10 questions.
For each question, a video and a baseline are drawn at random, and the participant compares our result with that of the baseline under the same instruction.
\Cref{fig:user_study_interface} shows the interface of one question.
In the first step, the participant reads the edit instruction and watches the source video to the end, after which the two edited videos appear.
In the second step, the two edited videos are shown side by side in random order as Video A and Video B, play synchronously on a shared timeline, and can be paused or enlarged.
The participant then answers three questions, which video carries out the instruction better, which video changes the parts that should not be edited less, and which video looks better in visual quality, choosing Video A, About the same or Video B for each.
The three choices are independent and can be revised until the participant confirms the question.
The participants are not told which video is ours.
In total, the study collects 420 comparisons, between 67 and 77 per baseline, and \Cref{fig:user_study} reports the percentages of comparisons in which the participants prefer the baseline, rate a tie or prefer our method.

In the example of \Cref{fig:user_study_interface}, the instruction asks to replace the group of cats with a single black cat near the front wheel of the truck, to change the weather to a foggy morning and to pan slowly to the right.
Video B, our result, replaces the cats with one black cat, turns the distant boats into soft silhouettes in the fog and pans the camera, whereas Video A keeps several cats and the sunny weather.

\makeatletter\setlength{\@fptop}{0pt}\makeatother

\newcommand{\arxivfilelist}[1]{}%
\arxivfilelist{%
\includegraphics{figs/analysis/complexity_editing_bars.pdf}%
\includegraphics{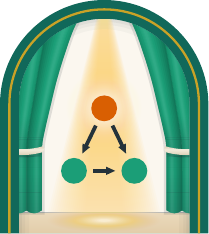}%
\includegraphics{figs/method/method_overview.pdf}%
\includegraphics{figs/oneword/0481_A.png}%
\includegraphics{figs/oneword/0481_B.png}%
\includegraphics{figs/qualitative/multiverse_chain.pdf}%
\includegraphics{figs/qualitative/qual_ablation_613.pdf}%
\includegraphics{figs/suppl/qual/edit_0303_bernini.jpg}%
\includegraphics{figs/suppl/qual/edit_0303_kling.jpg}%
\includegraphics{figs/suppl/qual/edit_0303_omniweaving.jpg}%
\includegraphics{figs/suppl/qual/edit_0303_ours.jpg}%
\includegraphics{figs/suppl/qual/edit_0303_seedance.jpg}%
\includegraphics{figs/suppl/qual/edit_0303_source.jpg}%
\includegraphics{figs/suppl/qual/edit_0351_kiwi.jpg}%
\includegraphics{figs/suppl/qual/edit_0351_kling.jpg}%
\includegraphics{figs/suppl/qual/edit_0351_omniweaving.jpg}%
\includegraphics{figs/suppl/qual/edit_0351_ours.jpg}%
\includegraphics{figs/suppl/qual/edit_0351_seedance.jpg}%
\includegraphics{figs/suppl/qual/edit_0351_source.jpg}%
\includegraphics{figs/suppl/qual/gen_leaf_cogvideox.jpg}%
\includegraphics{figs/suppl/qual/gen_leaf_dreamina.jpg}%
\includegraphics{figs/suppl/qual/gen_leaf_kling.jpg}%
\includegraphics{figs/suppl/qual/gen_leaf_ours.jpg}%
\includegraphics{figs/suppl/qual/gen_leaf_pixverse.jpg}%
\includegraphics{figs/suppl/qual/gen_monkey_cogvideox.jpg}%
\includegraphics{figs/suppl/qual/gen_monkey_dreamina.jpg}%
\includegraphics{figs/suppl/qual/gen_monkey_kling.jpg}%
\includegraphics{figs/suppl/qual/gen_monkey_ours.jpg}%
\includegraphics{figs/suppl/qual/gen_monkey_pixverse.jpg}%
\includegraphics{figs/suppl/qual/rev_0032_ditto.jpg}%
\includegraphics{figs/suppl/qual/rev_0032_kiwi.jpg}%
\includegraphics{figs/suppl/qual/rev_0032_lucy.jpg}%
\includegraphics{figs/suppl/qual/rev_0032_omniweaving.jpg}%
\includegraphics{figs/suppl/qual/rev_0032_ours.jpg}%
\includegraphics{figs/suppl/qual/rev_0032_source.jpg}%
\includegraphics{figs/suppl/qual/rev_0480_ditto.jpg}%
\includegraphics{figs/suppl/qual/rev_0480_kiwi.jpg}%
\includegraphics{figs/suppl/qual/rev_0480_lucy.jpg}%
\includegraphics{figs/suppl/qual/rev_0480_omniweaving.jpg}%
\includegraphics{figs/suppl/qual/rev_0480_ours.jpg}%
\includegraphics{figs/suppl/qual/rev_0480_source.jpg}%
\includegraphics{figs/suppl/user_study_interface.png}%
\includegraphics{figs/teaser/editing_bernini.jpg}%
\includegraphics{figs/teaser/editing_kling30omni.jpg}%
\includegraphics{figs/teaser/editing_omniweaving.jpg}%
\includegraphics{figs/teaser/editing_ours.jpg}%
\includegraphics{figs/teaser/editing_seedance.jpg}%
\includegraphics{figs/teaser/editing_source.jpg}%
\includegraphics{figs/teaser/generation_cogvideox.jpg}%
\includegraphics{figs/teaser/generation_dreamina.jpg}%
\includegraphics{figs/teaser/generation_kling.jpg}%
\includegraphics{figs/teaser/generation_pixverse.jpg}%
\includegraphics{figs/teaser/generation_v4a.jpg}%
\includegraphics{figs/teaser/revision_ditto.jpg}%
\includegraphics{figs/teaser/revision_kiwi.jpg}%
\includegraphics{figs/teaser/revision_lucy.jpg}%
\includegraphics{figs/teaser/revision_omniweaving.jpg}%
\includegraphics{figs/teaser/revision_ours.jpg}%
\includegraphics{figs/teaser/revision_source.jpg}%
\includegraphics{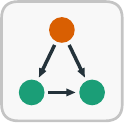}%
\includegraphics{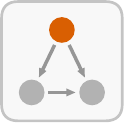}%
\includegraphics{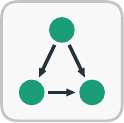}%
\includegraphics{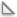}%
\includegraphics{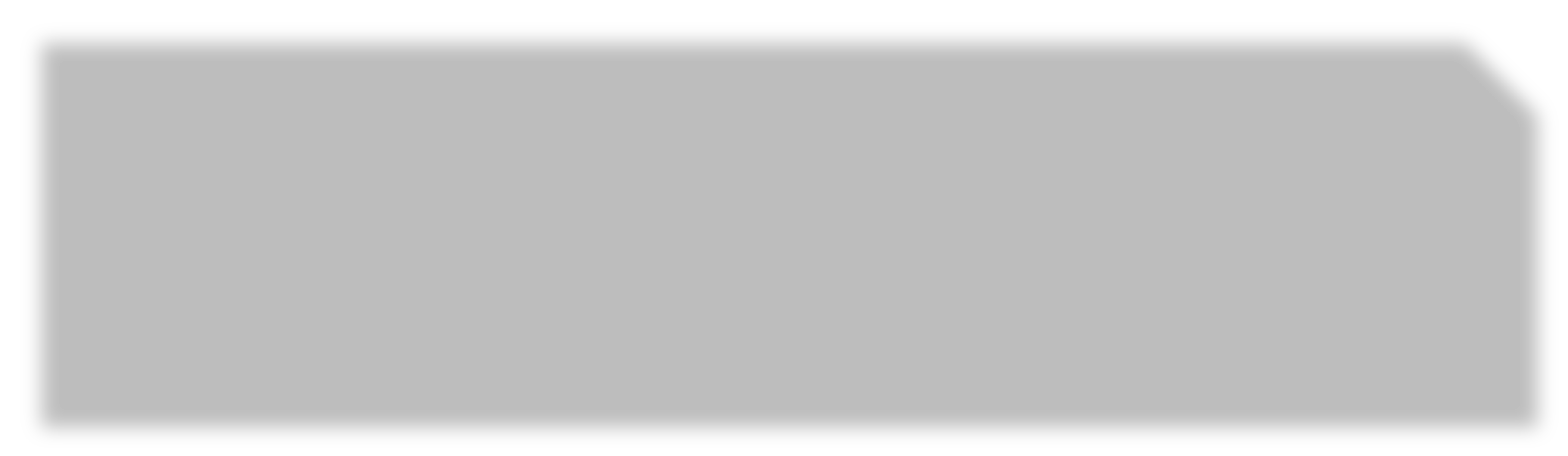}%
\includegraphics{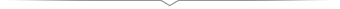}%
\includegraphics{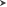}%
\includegraphics{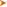}%
\includegraphics{figs/teaser_design/noise_strip.png}%
\includegraphics{figs/userstudy/user_study_full.pdf}%
}
\end{document}